\documentclass{article}
\usepackage{iclr2027_conference,times}

\usepackage{booktabs}
\usepackage{array}
\usepackage{graphicx}
\usepackage{wrapfig}
\usepackage{capt-of}
\graphicspath{{fig_file/}}
\usepackage{float}
\usepackage{placeins}
\usepackage{needspace}
\usepackage{xcolor}
\usepackage{colortbl}
\usepackage{multirow}
\usepackage{soul}
\usepackage{tcolorbox}

\definecolor{tabpanel}{RGB}{210,220,236}
\definecolor{tabtask}{RGB}{232,239,250}
\definecolor{tabdim}{RGB}{238,234,246}
\definecolor{tabmean}{RGB}{224,232,246}
\definecolor{tabours}{RGB}{220,236,250}
\usepackage{amsmath}
\usepackage{amssymb}
\usepackage{amsthm}
\usepackage{algorithm}
\usepackage{algorithmic}
\usepackage[T1]{fontenc}
\usepackage[utf8]{inputenc}
\usepackage{microtype}
\usepackage{inconsolata}
\usepackage{url}
\usepackage[hidelinks]{hyperref}

\title{EduAlign: Aligning Educational LLMs for Helpfulness, Personalization, and Creativity}
\author{%
\parbox[t]{\dimexpr\textwidth-2\tabcolsep\relax}{\bfseries\parindent=0pt
\mbox{Siyu Song\textsuperscript{2,3}},
\mbox{Ye Lu\textsuperscript{2,3}},
\mbox{Wentao Liu\textsuperscript{1,2,3}},
\mbox{Ruohua Zhang\textsuperscript{2,3}},
\mbox{Tao Liu\textsuperscript{2,3}},
\mbox{Jinze Lv\textsuperscript{2,3}},
\mbox{Xinyun Wang\textsuperscript{2,4}},
\mbox{Aimin Zhou\textsuperscript{1,2,3}},
\mbox{Fei Tan\textsuperscript{2}},
\mbox{Bo Jiang\textsuperscript{2}},
\mbox{Hao Hao\textsuperscript{2,}\thanks{Corresponding author.}}%
}\\[1em]
\textsuperscript{1}Shanghai Innavation Institute, Shanghai, 200231, China\\
\textsuperscript{2}Shanghai Institute of AI for Education, East China Normal University,\\
Shanghai, 200062, China\\
\textsuperscript{3}School of Computer Science and Technology,\\
East China Normal University, Shanghai, 200062, China\\
\textsuperscript{4}Department of Educational Information Technology,\\
East China Normal University, Shanghai, 200062, China
}
\iclrfinalcopy

\hypersetup{pdftitle={EduAlign: Aligning Educational LLMs for Helpfulness, Personalization, and Creativity},
  pdfauthor={Siyu Song, Ye Lu, Wentao Liu, Ruohua Zhang, Tao Liu, Jinze Lv, Xinyun Wang, Aimin Zhou, Fei Tan, Bo Jiang, Hao Hao}}

\begin{document}
\maketitle
\fancyhead{}
\renewcommand{\headrulewidth}{0pt}
\begin{abstract}
Educational language models should pursue three complementary objectives: helpfulness through responsible guidance, personalization to learner needs, and creativity that supports exploration. General-purpose models do not consistently satisfy these educational requirements, motivating explicit alignment with all three objectives. We present \textbf{EduAlign}, a framework that combines educational reward modeling with multi-objective reinforcement learning. To obtain training signals for these goals across educational scenarios, we develop a rubric-guided data synthesis pipeline that iteratively refines scoring criteria and generation prompts for each subject, grade band, and task type. The resulting supervision trains a multidimensional reward model, \textbf{Edu-RM}, to predict helpfulness, personalization, and creativity scores. To jointly optimize these educational objectives, we introduce Ideal-point Policy Optimization (\textbf{IdealPO}), a multi-objective reinforcement learning algorithm. \textbf{IdealPO} combines ideal-point marginal credit with continuous conflict weighting to guide policy updates that account for uneven progress and conflicting contributions across objectives. Experiments with 7B and 32B models show that \textbf{EduAlign} achieves the highest mean ratings on all three dimensions from both education experts and a model judge, alongside the highest aggregate scores on EduBench, ELMES, and EduValues among the compared methods. Reward-model validation and component ablations further support the effectiveness of the framework.
\end{abstract}

\section{Introduction}
Large language models (LLMs) are expanding the ways students access explanations, receive feedback, and explore ideas, while helping teachers develop learning materials and instructional activities. Educational systems such as EduChat and LearnLM illustrate this potential through specialized training for teaching and learning~\citep{dan2023educhat,learnlm2024improving}. Yet general-purpose models often prioritize delivering information or direct answers over guiding the learning process~\citep{sonkar2024pedagogical,learnlm2024improving}. Effective educational assistance also requires sensitivity to learners' needs and support for their own reasoning and exploration. These demands motivate educational alignment that explicitly translates educational goals into model behavior.

We organize this alignment around three objectives: \emph{Helpfulness}, \emph{Personalization}, and \emph{Creativity} (HPC). Helpfulness means honest, grounded, and responsible educational guidance. Personalization concerns adaptation to learners' stated needs and context; creativity concerns relevant, diverse, and actionable support for exploration. Aligning these objectives presents two challenges. First, their evaluation requires scoring standards that reflect different subjects, grade bands, and task types. EduBench and ELMES provide multidimensional educational assessments~\citep{xu2025edubench,wei2025elmesautomatedframeworkevaluating}, but translating shared HPC goals into consistent, scenario-specific reward supervision remains an open problem. Personalization, for example, can require age-appropriate language in an elementary-school explanation and guidance matched to prerequisite knowledge in a high-school exercise. Obtaining expert annotations across such varied settings is costly and time-consuming~\citep{sonkar2024pedagogical}. Second, existing educational alignment approaches emphasize pedagogical preferences or instruction following~\citep{sonkar2024pedagogical,learnlm2024improving}, leaving the joint optimization of HPC to be specified. The three objectives can progress unevenly, and a response can improve one while weakening another. A fixed mean reward gives each objective the same marginal weight regardless of its remaining deficit and obscures these opposing contributions. Reward-wise normalization, as in GDPO~\citep{liu2026gdpo}, addresses differences between reward distributions. Educational alignment additionally calls for response credits that account for objective deficits and conflicts. Together, these challenges require both scalable multidimensional reward modeling and an optimization method that uses the resulting reward vector.

To address these challenges, we propose \textbf{EduAlign}, a framework connecting educational reward modeling with multi-objective reinforcement learning. For reward modeling, we instantiate the shared HPC definitions as scenario-specific rubrics and jointly refine their scoring criteria and data-generation prompts. Response weaknesses guide prompt refinement, while score distributions and scoring rationales guide rubric refinement. The synthesized examples train \textbf{Edu-RM} to predict an HPC reward vector, providing training supervision without expert annotation of each generated response. For alignment, we introduce \emph{Ideal-point Policy Optimization (\textbf{IdealPO})}, combining ideal-point marginal credit (\textbf{IMC}) with continuous conflict weighting (\textbf{CCW}). \textbf{IMC} measures how removing a response changes the squared deficit between its rollout group's mean rewards and an ideal reward vector. This construction accounts for uneven progress across objectives when assigning response credits. \textbf{CCW} then attenuates credit magnitudes when dimensional contributions have opposing signs. The resulting credits are normalized within each prompt group to guide policy updates, connecting educational reward supervision to optimization that accounts for both uneven progress and conflicting contributions.

We evaluate \textbf{EduAlign} at 7B and 32B scales through expert and model-judge assessments and three public educational benchmarks. On 200 questions, education experts and Claude Sonnet 4.6 both assign \textbf{EduAlign} the highest mean rating in every HPC dimension at both scales. \textbf{EduAlign} also achieves the highest aggregate scores among the compared methods on EduBench, ELMES, and EduValues~\citep{zhang2025values}. Validation on 500 expert-labeled examples shows higher reward-model correlations after iterative synthesis, while 7B component ablations support the combined use of \textbf{IMC} and \textbf{CCW}. These findings demonstrate the effectiveness of \textbf{EduAlign} for improving the evaluated educational response qualities.

\section{Related Work}
\paragraph{Educational Alignment.}
Educational alignment adapts LLMs through pedagogical supervision, preference learning, and interaction-based rewards. Synthetic preferences encourage scaffolded guidance~\citep{sonkar2024pedagogical}, while LearnLM incorporates pedagogical instructions into supervised fine-tuning and RLHF~\citep{learnlm2024improving}. Rubric-based GPT-4 evaluations supply DPO preferences for mathematical feedback~\citep{scarlatos2024improving}. Subsequent work combines pedagogical rubric scores with predicted student correctness~\citep{scarlatos2025training}, and online RL with simulated students balances teaching quality and problem-solving accuracy~\citep{dinucu2025problem}. These approaches establish training signals for tutoring behavior and learner responses. \textbf{EduAlign} focuses on jointly aligning helpfulness, personalization, and creativity across educational scenarios through explicit rewards for each dimension.

\paragraph{Multi-Objective Reinforcement Learning.}
Multi-objective RL studies how to optimize potentially competing objectives~\citep{hayes2022practical}. In LLM training, reward normalization, uneven objective progress, and conflicting feedback all affect how multiple rewards guide policy updates. GRPO standardizes scalar rewards within prompt groups~\citep{shao2024deepseekmath}, whereas GDPO normalizes each reward dimension before aggregation~\citep{liu2026gdpo}. Focal Reward estimates rubric saturation through inverse reward projection~\citep{huang2026focal}, and SA-MRPO reweights normalized advantages using batch-level objective saturation~\citep{wang2026samrpo}. Both adapt optimization emphasis to the remaining room for improvement. GD$^2$PO addresses conflicting advantages through rollout filtering and query reweighting~\citep{liu2026gd2po}. \textbf{IdealPO} instead constructs response credits from leave-one-out changes in a prompt group's ideal-point loss and continuously attenuates credits with opposing dimensional contributions. This combines information about objective progress and conflict within the credit assigned to each response, guiding joint HPC alignment.

\section{\textbf{EduAlign}}
\label{sec:method}
\begin{figure}[t]
\centering
\includegraphics[width=0.90\linewidth]{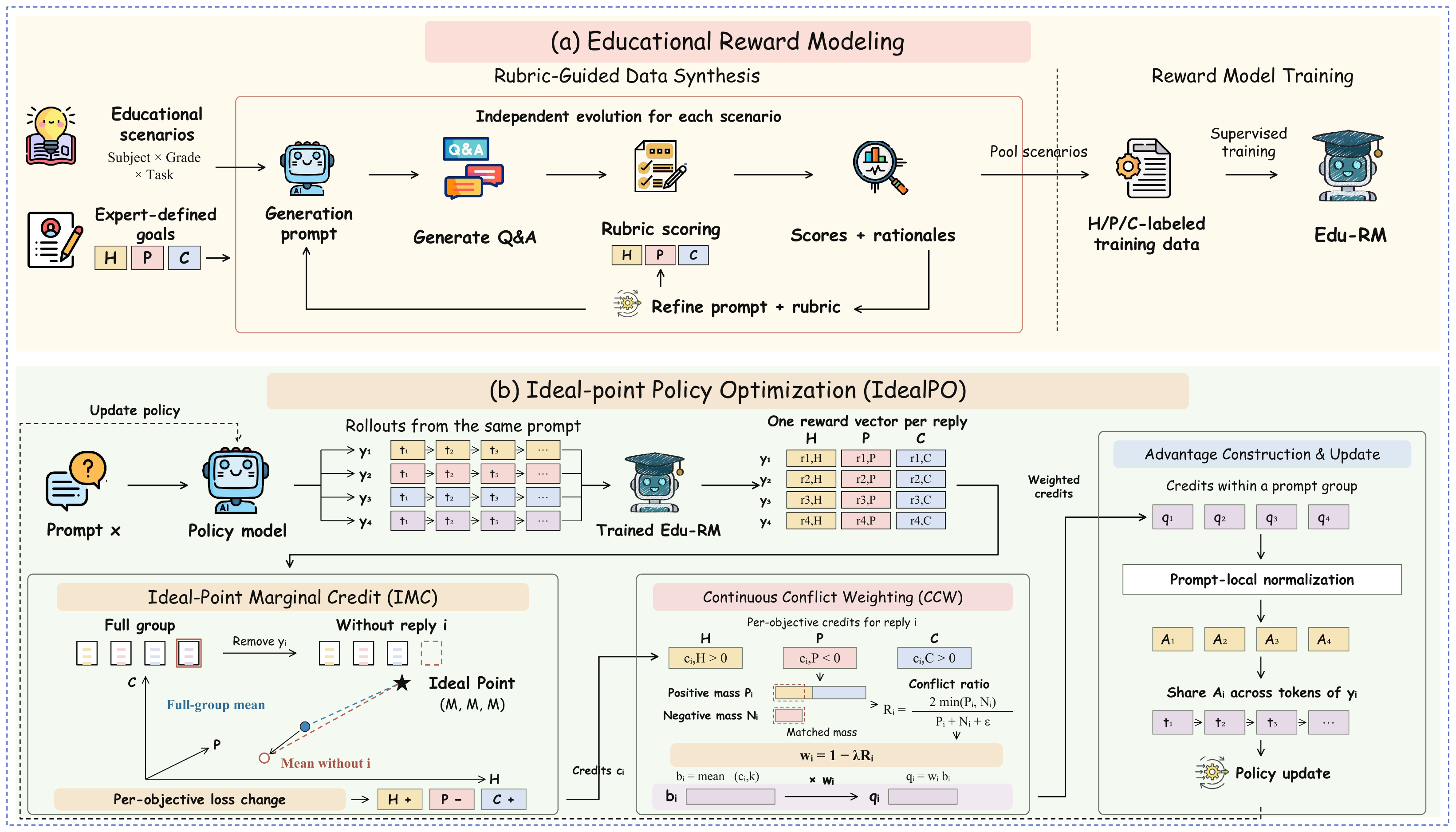}
\caption{Overview of \textbf{EduAlign}. (a) Expert-informed scenario rubrics guide data synthesis. Response weaknesses refine generation prompts; score distributions and rationales refine scoring standards. The resulting H/P/C labels train \textbf{Edu-RM} with a unified scoring rubric. (b) \textbf{IdealPO} uses \textbf{Edu-RM} scores to compute ideal-point marginal credits (\textbf{IMC}), attenuates credits with opposing dimensional contributions (\textbf{CCW}), and normalizes them within each prompt group for policy updates.}
\label{fig:framework}
\end{figure}

\textbf{EduAlign} connects the two challenges in Section~1: translating educational goals into scenario-specific reward supervision and using the resulting reward vector to guide joint optimization. As Figure~\ref{fig:framework} shows, rubric-guided synthesis trains \textbf{Edu-RM} to score helpfulness, personalization, and creativity (H/P/C). \textbf{IdealPO} then assigns response credits according to remaining objective deficits and conflicting contributions. Reward-model validation tests the quality of the supervision, while policy ablations test the credit construction.

\subsection{Educational Reward Modeling}
\label{sec:data}
Educational objectives must be shared across settings, while their scoring criteria must reflect the task and learner. We therefore separate the shared H/P/C definitions from their scenario-specific scoring standards. Iterative synthesis uses these standards to produce graded supervision, which \textbf{Edu-RM} learns through a common scoring interface.

\paragraph{From educational objectives to scenario rubrics.}
\emph{Helpfulness} covers honest, grounded, respectful, inclusive, and responsible guidance. \emph{Personalization} concerns adaptation to the learner's stated needs, context, and constraints, using conditional suggestions or clarification when information is missing. \emph{Creativity} concerns relevant new perspectives, diverse approaches, and thought-provoking guidance developed into actionable suggestions. Appendix~\ref{app:rm-prompts} gives the full definitions and score anchors.

We instantiate these objectives in 330 scenarios, each defined by $s=(\text{subject},\text{grade band},\text{task type})$. Educational criteria elicited through expert interviews initialize a rubric $\mathcal R_s^{(0)}$ with H/P/C indicators, scoring descriptions, and positive and negative examples. Scenario context, learner profiles, and generation requirements initialize a synthesis prompt $p_s^{(0)}$. This makes a shared objective concrete: personalization can require age-appropriate language in an elementary-school explanation and prerequisite-matched guidance in a high-school exercise.

\paragraph{Joint refinement of examples and scoring standards.}
A fixed generation prompt can repeatedly produce similar weaknesses, while ambiguous scoring descriptions can obscure differences in response quality. We use feedback from rubric assessments to refine both. At round $t$, GPT-5 generates question--response pairs $(x_{s,i}^{(t)},y_{s,i}^{(t)})$ from $p_s^{(t)}$, then scores them with $\mathcal R_s^{(t)}$. For each indicator $j$, it returns an integer score $a_{s,i,j}^{(t)}\in\{0,\ldots,5\}$ and a short rationale. Averaging over the indicators $\mathcal I_{s,k}$ assigned to objective $k$ yields
\begin{equation}
\widehat r_{s,i,k}^{(t)}=
\frac{1}{|\mathcal I_{s,k}|}\sum_{j\in\mathcal I_{s,k}}a_{s,i,j}^{(t)},
\qquad k\in\{H,P,C\}.
\label{eq:rubric}
\end{equation}
Response weaknesses and rationales guide the next generation prompt. Indicator-level score distributions and rationales guide clarification of scoring standards that poorly distinguish quality levels. GPT-5 performs both updates, keeping indicator identities and H/P/C assignments fixed. Each scenario maintains its own prompt--rubric state, so refinement responds to local educational requirements.

The loop runs for at most five rounds per scenario. A batch score above the historical best retains its examples and prompt--rubric state; a lower or tied score restores the best state and stops iteration. Two fixed question--response anchors per scenario monitor changes in rubric scoring and trigger drift warnings. Appendix~\ref{app:synthesis-algorithm} specifies this selection procedure and the role of anchors; Appendix~\ref{app:synthesis-prompts} provides all synthesis and refinement prompts. Pooling accepted examples gives 12,390 QA pairs with graded H/P/C labels.

\paragraph{Learning and validating a common reward interface.}
Scenario rubrics supply the training labels, while a unified H/P/C rubric serves as the scoring instruction for every \textbf{Edu-RM} input. The question and response supply the context needed to apply these criteria. We combine the synthesized corpus with 15,000 scored low-quality responses, giving 27,390 training examples. Dimensional averages are rounded to integers, with half-integer ties rounded upward, and serialized into a structured score sequence. SFT trains \textbf{Edu-RM} to predict that sequence from the QA pair and unified rubric. At rollout time, the same interface returns $f_\phi(x,y)=(r_H,r_P,r_C)$. Appendix~\ref{app:rm-prompts} gives the input--output format; Appendix~\ref{app:training} gives the SFT objective, negative-example construction, and training settings.

The synthesis loop selects data using its own rubric scores. We assess the learned scorer against expert judgments in Section~\ref{sec:rm-validation}: reward models trained before and after refinement use matched corpus sizes, the same negative examples, and identical training settings. Both are evaluated on the same 500 expert-labeled pairs whose questions are excluded from RM training. Higher Pearson and Spearman correlations in all three dimensions support the use of iterative synthesis for expert-aligned reward supervision.

\subsection{Ideal-point Policy Optimization}
\label{sec:optimization}
A fixed reward mean assigns the same marginal value to every dimension even when their remaining deficits differ. It also collapses opposing dimensional contributions into a single net score. \textbf{IdealPO} addresses these two properties in sequence: \textbf{IMC} assigns credit relative to the group's distance from the ideal reward vector, and \textbf{CCW} moderates credits with conflicting contributions before group normalization.

For a prompt $x$, we sample $n>1$ responses from $\pi_{\mathrm{old}}$ and score each independently with \textbf{Edu-RM}, obtaining $\mathbf r_i\in[0,M]^K$. Here $K=3$, $M=5$, and $n=8$ in our experiments. All subsequent credit calculations use this prompt's response group.

\paragraph{Ideal-Point Marginal Credit (IMC).}
\label{sec:loo}
The ideal vector $\mathbf r^\star=M\mathbf1$ represents the score ceiling on all objectives. We measure the squared deficit of the group mean $\mu_k=n^{-1}\sum_i r_{i,k}$ from this target:
\begin{equation}
\mathcal L_{\mathrm{ideal}}(\boldsymbol\mu)
=\frac1K\sum_{k=1}^K\ell_k,
\qquad \ell_k=\left(1-\frac{\mu_k}{M}\right)^2.
\label{eq:ideal}
\end{equation}
For response $i$, the leave-one-out mean is $\mu_{-i,k}=(n\mu_k-r_{i,k})/(n-1)$. Its dimensional credit and scalar base credit are
\begin{equation}
c_{i,k}=\left(1-\frac{\mu_{-i,k}}{M}\right)^2-\ell_k,
\qquad b_i=\frac1K\sum_k c_{i,k}.
\label{eq:credit}
\end{equation}
A positive $c_{i,k}$ means that removing the response increases the group's deficit on objective $k$; a negative value means removal reduces it. Thus $b_i$ measures the response's net contribution to reducing the ideal-point loss relative to its sampled peers.

Let $\delta_{i,k}=r_{i,k}-\mu_k$ and $d_k=1-\mu_k/M$. Expanding Equation~\eqref{eq:credit} gives
\begin{equation}
c_{i,k}=\frac{2d_k\delta_{i,k}}{M(n-1)}
+\frac{\delta_{i,k}^2}{M^2(n-1)^2}.
\label{eq:expansion}
\end{equation}
For a fixed positive reward deviation, the credit increases with the remaining deficit $d_k$. Objectives farther below the ceiling therefore receive greater marginal emphasis within the current group. The second term is the finite-group quadratic correction. Appendix~\ref{app:credit-properties} derives this expression and its sign properties. Section~\ref{sec:ablation} evaluates this credit construction against a conflict-weighted raw-mean alternative.

\paragraph{Continuous Conflict Weighting (CCW).}
\label{sec:conflict}
Deficit-sensitive credits can still combine gains in personalization or creativity with a loss in helpfulness. Their average $b_i$ gives the net contribution, but equal net credits can contain different amounts of opposing contributions. We quantify this opposition through positive and negative credit masses:
\begin{equation}
P_i=\frac1K\sum_k\max(c_{i,k},0),\qquad
N_i=\frac1K\sum_k\max(-c_{i,k},0).
\end{equation}
Since $b_i=P_i-N_i$, the smaller mass measures the part offset by contributions of the other sign. We define the conflict ratio $R_i$ and weighted credit $q_i$ as
\begin{equation}
R_i=\frac{2\min(P_i,N_i)}{P_i+N_i+\varepsilon},\qquad
w_i=1-\lambda R_i,\qquad q_i=w_i b_i.
\label{eq:conflict}
\end{equation}
Contributions with a common sign have $R_i=0$. Opposing signs give $R_i>0$, with stronger attenuation as positive and negative masses become more balanced. We use $\lambda=0.5$ and $\varepsilon=10^{-8}$, so $w_i\in[0.5,1]$. CCW moves both positive and negative conflicting credits toward zero while preserving their signs before normalization. Its effect on the final advantages depends on the other responses in the group; Appendix~\ref{app:credit-properties} explains this distinction. The $\lambda=0$ ablation in Section~\ref{sec:ablation} tests its contribution to benchmark performance, with task-level results in Appendix~\ref{app:ablation-details}.

\paragraph{From response credits to policy updates.}
We standardize $q_i$ using its prompt-group mean $\bar q$ and population standard deviation $\sigma_q$:
\begin{equation}
A_i=\begin{cases}(q_i-\bar q)/\sigma_q,&\sigma_q>\varepsilon,\\0,&\text{otherwise.}\end{cases}
\label{eq:advantage}
\end{equation}
The resulting advantage expresses each weighted credit relative to its group. It is shared across the response's tokens and used in a clipped policy objective with a reference-policy KL penalty~\citep{schulman2017proximal,shao2024deepseekmath}. We aggregate over valid response tokens and apply no second batch-level advantage normalization. Appendix~\ref{app:policy-objective} specifies the objective, Algorithm~\ref{alg:loo} gives the full procedure, and Appendix~\ref{app:config} lists training parameters. Section~\ref{sec:reward-curves} tracks the three reward channels during training.

\section{Experiments}
\label{sec:experiment}
Our experiments evaluate both stages of \textbf{EduAlign}. Expert and model-judge ratings test joint H/P/C alignment, and public benchmarks assess educational task performance. We then test whether iterative synthesis improves reward supervision, whether both \textbf{IMC} and \textbf{CCW} contribute to policy performance, and how the three reward channels evolve during training.

\subsection{Experimental Setup}
\paragraph{Policies and comparisons.}
The policy backbones are Qwen2.5-7B-Instruct and Qwen2.5-32B-Instruct~\citep{qwen2025qwen25technicalreport}. Comparisons include the instruction-tuned models, supervised fine-tuning (SFT), PPO~\citep{schulman2017proximal}, DAPO~\citep{yu2025dapo}, GRPO~\citep{shao2024deepseekmath}, and GDPO~\citep{liu2026gdpo}. GDPO directly tests whether our credit construction offers benefits beyond decoupled reward normalization. GRPO, GDPO, \textbf{EduAlign}, and the two ablations are trained for 40 rollout steps; public-benchmark results use the final checkpoint.

\paragraph{\textbf{EduAlign} configuration.}
The RL training set contains 3,880 synthesized prompts covering educational and general questions. Policy SFT uses 5,000 question--response pairs selected from the iteratively synthesized data based on their combined helpfulness, personalization, and creativity scores. Each rollout round samples 128 prompts and eight responses per prompt, yielding 1,024 responses. The global update batch is 512, with two updates per rollout round, totaling 80 parameter updates over 40 rounds. Both model scales use a learning rate of $10^{-6}$, temperature 1.0, conflict strength 0.5, clipping parameter 0.2, and KL coefficient 0.001. The 7B and 32B runs use four and eight GPUs, respectively. Appendix~\ref{app:config} records the remaining supplied configuration. GRPO and GDPO use the same rollout and optimization settings as \textbf{EduAlign}.

\paragraph{Public benchmarks.}
We select tasks relevant to helpfulness, personalization, and creativity. From EduBench~\citep{xu2025edubench}, we use proportional random sampling to select 100 questions per task for Idea Provision (IP), Personalized Learning Support (PLS), and Teaching Material Generation (TMG). From ELMES~\citep{wei2025elmesautomatedframeworkevaluating}, we use all questions for Interdisciplinary Lesson Plan Generation (ILPG), Contextual Question Generation (CQG), and Knowledge Point Explanation (KPE). AVG is the arithmetic mean of the three task scores. Appendix~\ref{app:fullresults} reports task-specific rubric results. For EduValues~\citep{zhang2025values}, we evaluate on all its Subjective Analysis (SA) questions and report the average score on a 1--5 scale. All three benchmarks use the scoring prompts from their original papers. We use Qwen3.5-397B-A17B to score EduValues and ELMES responses. For EduBench, we use DeepSeek V3, following the original study's finding of high agreement with human ratings~\citep{xu2025edubench}.

\subsection{Dimension-wise Alignment: Expert and Model-Judge Evaluation}
\label{sec:human}
\paragraph{Evaluation protocol.}
We collected questions through interviews with K--12 students and selected 200 for evaluation. Appendix~\ref{app:human} reports their category distribution, with 164 education-related questions (82\%) and 36 general questions (18\%). Five annotators, comprising researchers and doctoral students working on intelligent education, rate the responses. Each question is assigned to two experts; the same pair rates the responses from all compared models for that question. Raters see the question and response without model identity. We average the two expert ratings for each response and dimension, then average across questions. The overall score is the mean of the three dimension scores. Claude Sonnet 4.6 scores the same 200 questions and the same generated responses using the same 0--5 rubric, with model identities hidden. Responses are not regenerated between evaluators.

Table~\ref{tab:human} reports all three dimensions for each evaluator. We display them in HPC order: educational helpfulness (H), personalization (P), and creativity (C). Appendix~\ref{app:human} summarizes the question distribution.

\begin{table}[H]
\centering
\caption{Matched expert and Claude Sonnet 4.6 ratings on 200 questions ($\uparrow$, 0--5). Both evaluate identical responses with the same rubric and hidden model identities; expert scores average two raters per response. H/P/C are helpfulness, personalization, and creativity. Bold marks the best score within each evaluator and model scale. The + rows are separate training comparisons, not successive additions.}
\label{tab:human}
\small
\setlength{\tabcolsep}{3.5pt}
\renewcommand{\arraystretch}{1.12}
\setlength{\aboverulesep}{0.35pt}
\setlength{\belowrulesep}{0.35pt}
\begin{tabular}{llrrrrrrrr}
\toprule
&&\multicolumn{4}{c}{Education experts}&\multicolumn{4}{c}{Claude Sonnet 4.6}\\
\cmidrule(lr){3-6}\cmidrule(lr){7-10}
Scale & Method & H & P & C & Mean & H & P & C & Mean\\
\midrule
7B & Instruct &4.5775&3.5950&3.1225&3.7650&4.5750&3.4750&2.9750&3.6750\\
& +GRPO &4.7250&4.2600&3.9525&4.3125&4.7800&3.8650&3.8750&4.1733\\
& +GDPO &4.6850&4.2225&3.8350&4.2475&4.7600&3.8300&3.7350&4.1083\\
\rowcolor{tabours}
& \textbf{+EduAlign} &\textbf{4.7925}&\textbf{4.3225}&\textbf{4.0875}&\textbf{4.4008}&\textbf{4.7975}&\textbf{3.9225}&\textbf{4.0225}&\textbf{4.2475}\\
\midrule
32B & Instruct &4.5975&3.7550&3.1775&3.8433&4.6450&3.5200&2.9950&3.7200\\
& +GRPO &4.8500&4.3375&4.1325&4.4400&4.8400&4.0200&4.0200&4.2933\\
& +GDPO &4.8250&4.3175&4.1150&4.4192&4.8500&3.9850&4.0150&4.2833\\
\rowcolor{tabours}
& \textbf{+EduAlign} &\textbf{4.8850}&\textbf{4.5150}&\textbf{4.2850}&\textbf{4.5617}&\textbf{4.8825}&\textbf{4.0750}&\textbf{4.1350}&\textbf{4.3642}\\
\bottomrule
\end{tabular}
\end{table}

\paragraph{Expert-rated improvements across all three dimensions.}
\textbf{EduAlign} receives the highest expert scores on H, P, and C at both scales. Relative to GRPO, the strongest expert-rated alternative in every dimension, its 7B gains are 0.0675, 0.0625, and 0.1350 in HPC order; the corresponding 32B gains are 0.0350, 0.1775, and 0.1525. Its mean exceeds GRPO by 0.0883 at 7B and 0.1217 at 32B, and GDPO by 0.1533 and 0.1425. These dimension-wise results show that the higher expert-rated mean accompanies gains in each measured objective, rather than compensating for a lower score in another.

\paragraph{Corroborating model-judge assessment.}
Claude Sonnet 4.6 also assigns \textbf{EduAlign} the highest score in every dimension at both scales, with means of 4.2475 and 4.3642. Against the strongest alternative in each dimension, the 7B margins are 0.0175, 0.0575, and 0.1475 in HPC order; the 32B margins are 0.0325, 0.0550, and 0.1150. The two evaluators therefore give consistent evidence of gains across the three rated dimensions. Their absolute scores differ: for example, 32B \textbf{EduAlign} receives 4.5150 for personalization from experts and 4.0750 from Claude. Consistent model-level gains do not imply interchangeable scores or sample-level agreement. These are descriptive comparisons of mean response-quality ratings, without statistical-significance or student-learning claims. Appendix~\ref{app:qualitative} presents paired response excerpts illustrating the three dimensions.

\FloatBarrier
\subsection{Performance across Public Benchmarks}

Table~\ref{tab:main} compares all three benchmarks in separate 7B and 32B row blocks. The + rows denote separate training comparisons from the corresponding Instruct model. Comparisons are made within each benchmark because their scales differ. Appendix~\ref{app:fullresults} provides task-specific rubric scores for the compared methods. Reported differences are descriptive checkpoint comparisons, without repeated-run distributions or significance claims.

\begin{table}[!htbp]
\centering
\caption{Public-benchmark performance ($\uparrow$). AVG averages the three task scores within each benchmark. Bold marks the best score within each model scale.}
\label{tab:main}
\small
\setlength{\tabcolsep}{3pt}
\renewcommand{\arraystretch}{1.12}
\setlength{\aboverulesep}{0.35pt}
\setlength{\belowrulesep}{0.35pt}
\begin{tabular}{@{}l*{9}{c}@{}}
\toprule
& \multicolumn{4}{c}{\textbf{EduBench}} & \multicolumn{4}{c}{\textbf{ELMES}} & \textbf{EduValues}\\
\cmidrule(lr){2-5}\cmidrule(lr){6-9}\cmidrule(l){10-10}
Model & IP & PLS & TMG & AVG & ILPG & CQG & KPE & AVG & Score\\
\midrule
\textbf{7B-Instruct} & 6.867 & 8.510 & 8.901 & 8.093 & 3.313 & 3.240 & 3.100 & 3.218 & 3.575\\
\quad+SFT & 6.740 & 8.530 & 8.996 & 8.089 & 3.800 & 3.440 & 3.333 & 3.524 & 3.664\\
\quad+PPO & 7.400 & 8.694 & 8.960 & 8.351 & 3.700 & 3.620 & 2.500 & 3.273 & 3.717\\
\quad+DAPO & 7.819 & 8.868 & 9.041 & 8.576 & 4.010 & 3.360 & 3.380 & 3.583 & 3.876\\
\quad+GRPO & 7.834 & 9.028 & 9.209 & 8.690 & 4.107 & 3.660 & 3.433 & 3.733 & 4.027\\
\quad+GDPO & 7.845 & 8.998 & 9.160 & 8.668 & 3.980 & 3.660 & \textbf{3.483} & 3.708 & 4.009\\
\rowcolor{tabours}
\quad\textbf{+EduAlign} & \textbf{8.245} & \textbf{9.078} & \textbf{9.230} & \textbf{8.851} & \textbf{4.287} & \textbf{4.000} & 3.417 & \textbf{3.901} & \textbf{4.097}\\
\midrule
\textbf{32B-Instruct} & 7.985 & 8.366 & 8.971 & 8.441 & 3.520 & 3.680 & 3.633 & 3.611 & 3.788\\
\quad+SFT & 8.094 & 8.994 & 9.046 & 8.711 & 4.100 & 4.040 & 3.217 & 3.786 & 3.814\\
\quad+PPO & 7.820 & 8.954 & 9.147 & 8.640 & 3.910 & 4.220 & 3.120 & 3.750 & 3.903\\
\quad+DAPO & 8.321 & 9.116 & 9.230 & 8.889 & 4.327 & 3.960 & 3.800 & 4.029 & 4.124\\
\quad+GRPO & 8.234 & 9.080 & 9.241 & 8.852 & 4.487 & 4.100 & 3.733 & 4.107 & 4.133\\
\quad+GDPO & 8.275 & 9.116 & 9.251 & 8.881 & 4.433 & 4.160 & 3.633 & 4.075 & 4.142\\
\rowcolor{tabours}
\quad\textbf{+EduAlign} & \textbf{8.547} & \textbf{9.172} & \textbf{9.281} & \textbf{9.000} & \textbf{4.513} & \textbf{4.640} & \textbf{3.933} & \textbf{4.362} & \textbf{4.177}\\
\bottomrule
\end{tabular}
\end{table}

Table~\ref{tab:main} shows aggregate improvements at both scales. At 7B, \textbf{EduAlign} exceeds the strongest alternative by 0.161 on EduBench AVG, 0.168 on ELMES AVG, and 0.070 on EduValues. At 32B, the differences are 0.111, 0.255, and 0.035, respectively. Among the 32B baselines, DAPO leads on EduBench, GRPO on ELMES, and GDPO on EduValues. Relative to GDPO specifically, \textbf{EduAlign} improves the 7B/32B scores by 0.183/0.119 on EduBench AVG, 0.194/0.287 on ELMES AVG, and 0.088/0.035 on EduValues.

The disaggregated scores reveal limits to these gains. On ELMES, 7B \textbf{EduAlign} obtains 3.417 on KPE, below GRPO's 3.433 and GDPO's 3.483. The task-specific results in Appendix~\ref{app:fullresults} also show that DAPO exceeds \textbf{EduAlign} on higher-order thinking for idea provision at both scales (7.470 versus 7.360 at 7B; 7.870 versus 7.590 at 32B). Thus the results support stronger aggregate educational performance across benchmark definitions, but not Pareto dominance. They do not establish generalization to unseen student populations or improvements in actual learning outcomes.

\subsection{Reward Model Validation}
\label{sec:rm-validation}
We test whether jointly refining generation prompts and scoring standards produces more effective supervision for \textbf{Edu-RM}. This requires agreement with expert judgments: increasing the synthesis judge's own scores would not establish that the learned reward signal better reflects educational quality.

\paragraph{Comparison at matched data volume.}
The before and after checkpoints each use 12,390 synthesized examples, from iter~0 and iterated synthesis, respectively, together with the same 15,000 negative examples and identical training settings. Both are evaluated on the same 500 QA pairs sampled from the final synthesized pool, with their questions excluded from RM training. All ten experts score every pair; the modal rating supplies the reference for each H/P/C dimension. Pearson $r$ measures linear association with expert scores, and Spearman $\rho$ measures agreement in response ordering. Appendix~\ref{app:training} gives the training protocol.

\paragraph{More informative supervision across H/P/C.}
Table~\ref{tab:rm-agree} shows improvements in both measures for every objective. Pearson $r$ increases by 0.1395, 0.1456, and 0.1817 for H/P/C, while Spearman $\rho$ increases by 0.1537, 0.1521, and 0.1952. Creativity benefits most, reaching $r=0.8937$ and $\rho=0.8949$. The consistent gains in association and ranking show that the refined data teach the reward model to distinguish response quality more closely in line with experts. Because data volume and training settings are held fixed, this comparison supports the effectiveness of the iterative synthesis procedure as a source of reward supervision. It evaluates joint prompt--rubric refinement on held-out questions from the synthesized distribution; the separate effects of the two updates are not isolated.

\subsection{Component Ablations}
\label{sec:ablation}
\begin{figure}[tb]
\centering
\begin{minipage}[t]{0.45\textwidth}
\vspace{0pt}
\centering
\captionof{table}{\textbf{Edu-RM}--expert correlations on the same 500 examples ($\uparrow$). Bold marks the higher value.}
\label{tab:rm-agree}
\small
\setlength{\tabcolsep}{3pt}
\renewcommand{\arraystretch}{1.12}
\begin{tabular}{@{}lrrrr@{}}
\toprule
& \multicolumn{2}{c}{Pearson $r$} & \multicolumn{2}{c}{Spearman $\rho$}\\
\cmidrule(lr){2-3}\cmidrule(lr){4-5}
& Before & After & Before & After\\
\midrule
H & 0.6207 & \textbf{0.7602} & 0.5139 & \textbf{0.6676}\\
P & 0.6907 & \textbf{0.8363} & 0.6782 & \textbf{0.8303}\\
C & 0.7120 & \textbf{0.8937} & 0.6997 & \textbf{0.8949}\\
\bottomrule
\end{tabular}
\end{minipage}\hfill
\begin{minipage}[t]{0.52\textwidth}
\vspace{0pt}
\centering
\includegraphics[width=0.80\linewidth]{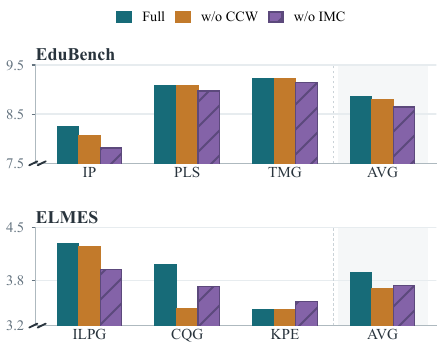}
\caption{Ablation results on EduBench and ELMES (7B).}
\label{fig:ablation}
\end{minipage}
\end{figure}

Figure~\ref{fig:ablation} compares three task scores and their arithmetic mean (AVG) for \textbf{EduAlign} and two 7B component ablations. Both retain the remaining training settings of the full method.

\emph{w/o \textbf{CCW}} sets $\lambda=0$ while retaining \textbf{IMC} and prompt-local normalization. Removing conflict weighting reduces the three-task average from 8.851 to 8.788 on EduBench and from 3.901 to 3.692 on ELMES. The largest task-level ELMES difference is on CQG (4.000 versus 3.420). The ELMES creativity score falls from 4.075 to 3.600, while personalization is slightly higher without conflict weighting (4.175 versus 4.150). Conflict weighting therefore changes the objective profile rather than uniformly increasing every score.

\emph{w/o \textbf{IMC}} uses within-group centered raw rewards $z_{i,k}=r_{i,k}-\mu_k$ in place of \textbf{IMC} contributions to estimate objective conflict, applies the resulting \textbf{CCW} weight to each response's raw reward averaged across H/P/C, and then performs within-group normalization. The ablation obtains three-task averages of 8.644 on EduBench and 3.727 on ELMES, below the full method's 8.851 and 3.901, respectively. EduBench IP declines from 8.245 to 7.811, although ELMES KPE is higher without \textbf{IMC} (3.520 versus 3.417). Both removals lower benchmark averages, supporting the combined credit construction with task-level trade-offs. Appendix~\ref{app:ablation-details} gives the full 7B differences.

\FloatBarrier
\subsection{Training Dynamics}
\label{sec:reward-curves}
\begin{figure}[!t]
\centering
\includegraphics[width=\linewidth]{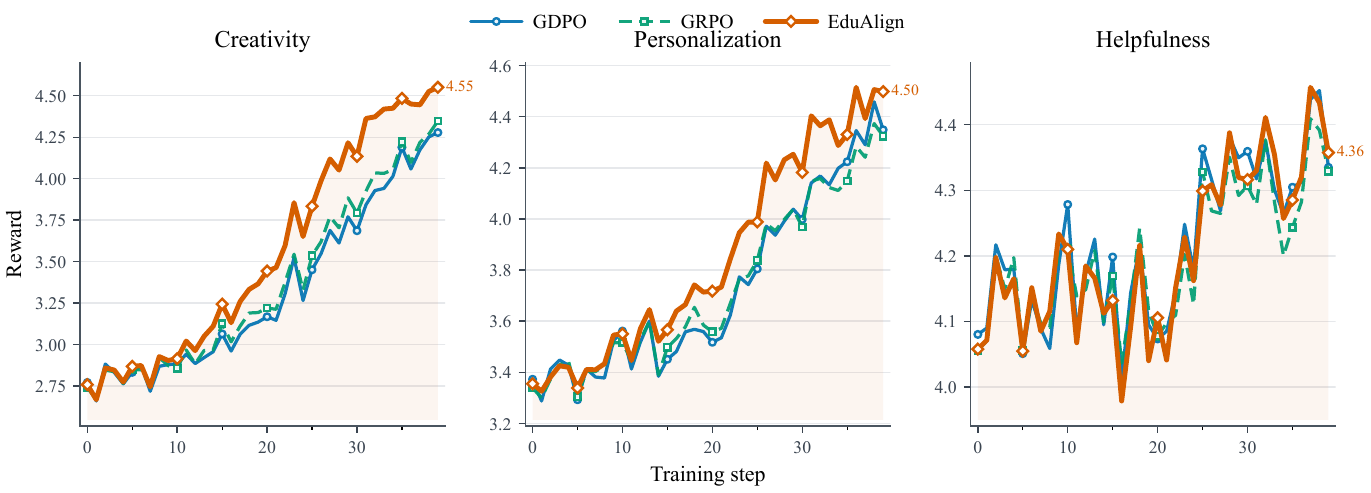}
\caption{Unsmoothed 7B rollout rewards over steps 0--39. \textbf{EduAlign} uses \textbf{IMC} and \textbf{CCW} with $\lambda=0.5$.}
\label{fig:reward-curves}
\end{figure}

Figure~\ref{fig:reward-curves} tracks unsmoothed \textbf{Edu-RM} rollout rewards over 40 steps of 7B training. The runs start at similar reward levels: creativity is about 2.75, personalization 3.35, and helpfulness 4.06. The larger initial deficits in creativity and personalization make them informative tests of the motivation for deficit-sensitive credit.

\paragraph{Progress on initially weaker objectives.}
\textbf{EduAlign} separates from GRPO and GDPO on creativity and personalization as training proceeds, reaching 4.551 and 4.498 at step~39. The strongest baseline endpoint is 4.348 on creativity and 4.349 on personalization. This advantage persists across the final ten steps: \textbf{EduAlign}'s mean rewards exceed the stronger baseline in each dimension by 0.318 and 0.156, respectively. The sustained gains are consistent with \textbf{IMC}'s greater marginal emphasis on objectives with larger remaining deficits, as expressed in Equation~\eqref{eq:expansion}.

\paragraph{Maintaining the stronger reward channel.}
Helpfulness fluctuates across rollout batches and remains close across methods. Over the final ten steps, its mean is 4.352 for \textbf{EduAlign}, compared with 4.346 for GDPO and 4.310 for GRPO. Thus, the larger gains in creativity and personalization accompany a comparable helpfulness level. These single-run trajectories characterize optimization under the learned scorer; the component ablations test \textbf{CCW}'s contribution, while expert ratings and public benchmarks assess response quality beyond training rewards.

\section{Conclusion}
\textbf{EduAlign} aligns educational LLMs with helpfulness, personalization, and creativity through expert-informed reward supervision and multi-objective policy optimization. Starting from rubrics generated from expert interviews, iterative data synthesis jointly refines generation prompts and scoring standards to construct discriminative supervision for \textbf{Edu-RM}. \textbf{IdealPO} promotes balanced improvement by emphasizing objectives with larger remaining deficits and moderately downweighting conflicting response credits. At both 7B and 32B scales, \textbf{EduAlign} achieves the highest aggregate scores among the compared methods on EduBench, ELMES, and EduValues, and the highest expert and Claude Sonnet 4.6 ratings in each H/P/C dimension. Reward-model validation shows closer agreement with expert judgments after iterative synthesis, and ablations support the combined use of \textbf{IMC} and \textbf{CCW}. These aggregate gains coexist with trade-offs on individual tasks. Future work will test whether the improved response quality leads to better learning outcomes across diverse learners and classroom settings.

\bibliography{custom}

@inproceedings{liu2026gdpo,
  title={{GDPO}: Group reward-Decoupled Normalization Policy Optimization for Multi-reward {RL} Optimization},
  author={Liu, Shih-Yang and Dong, Xin and Lu, Ximing and Diao, Shizhe and Belcak, Peter and Liu, Mingjie and Chen, Min-Hung and Yin, Hongxu and Wang, Yu-Chiang Frank and Cheng, Kwang-Ting and Choi, Yejin and Kautz, Jan and Molchanov, Pavlo},
  booktitle={Proceedings of the 43rd International Conference on Machine Learning},
  year={2026},
  url={https://research.nvidia.com/labs/twn/publication/icml_2026_gdpo/}
}

@article{liu2026gd2po,
  title={{GD$^2$PO}: Mitigating Multi-Reward Conflicts via Group-Dynamic reward-Decoupled Policy Optimization},
  author={Liu, Haotian and Liu, Yihao and Ni, Jingwei and Huang, Siyuan and Liu, Xinpeng and Cheng, Pengyu and Song, Jiajun and Ding, Ruijin and Li, Junfeng and Yu, Zhechao and Zhou, Mengyu and Xu, Hongteng and Jiang, Xiaoxi and Jiang, Guanjun},
  journal={arXiv preprint arXiv:2606.16771},
  year={2026},
  url={https://arxiv.org/abs/2606.16771}
}

@article{wang2026samrpo,
  title={Learn What's Left, Not What's Mastered: Saturation Aware Advantage Reweighting for Multi-Reward Policy Optimization},
  author={Wang, Yixuan and Chen, Yifei and Zhang, Haichao and Luo, Haozheng and Wu, Xander and Ni, Jie and Fu, Yun and Vasconcelos, Nuno and Li, Yijiang},
  journal={arXiv preprint arXiv:2608.16072},
  year={2026},
  url={https://arxiv.org/abs/2608.16072}
}

@article{huang2026focal,
  title={Focal Reward: Balanced Reinforcement Learning under Rubric-Based Rewards},
  author={Huang, Yu and Zhao, Zihua and Huan, Zhaoxin and Gu, Wanli and Hong, Feng and Ge, Xinmu and Yuan, Lin and Wu, Weichang and Hu, Qiang and Zhang, Xiaolu and Zhou, Jun and Yao, Jiangchao},
  journal={arXiv preprint arXiv:2605.26579},
  year={2026},
  url={https://arxiv.org/abs/2605.26579}
}

@article{yu2025dapo,
  title={Dapo: An open-source llm reinforcement learning system at scale},
  author={Yu, Qiying and Zhang, Zheng and Zhu, Ruofei and Yuan, Yufeng and Zuo, Xiaochen and Yue, Yu and Dai, Weinan and Fan, Tiantian and Liu, Gaohong and Liu, Lingjun and others},
  journal={arXiv preprint arXiv:2503.14476},
  year={2025}
}

@inproceedings{scarlatos2025training,
  title={Training {LLM}-Based Tutors to Improve Student Learning Outcomes in Dialogues},
  author={Scarlatos, Alexander and Liu, Naiming and Lee, Jaewook and Baraniuk, Richard and Lan, Andrew},
  booktitle={Artificial Intelligence in Education},
  series={Lecture Notes in Computer Science},
  volume={15877},
  pages={251--266},
  publisher={Springer},
  year={2025},
  doi={10.1007/978-3-031-98414-3_18},
  url={https://link.springer.com/chapter/10.1007/978-3-031-98414-3_18}
}

@inproceedings{dinucu2025problem,
  title={From Problem-Solving to Teaching Problem-Solving: Aligning {LLMs} with Pedagogy using Reinforcement Learning},
  author={Dinucu-Jianu, David and Macina, Jakub and Daheim, Nico and Hakimi, Ido and Gurevych, Iryna and Sachan, Mrinmaya},
  booktitle={Proceedings of the 2025 Conference on Empirical Methods in Natural Language Processing},
  pages={272--292},
  publisher={Association for Computational Linguistics},
  year={2025},
  doi={10.18653/v1/2025.emnlp-main.15},
  url={https://aclanthology.org/2025.emnlp-main.15/}
}

@inproceedings{scarlatos2024improving,
  title={Improving the validity of automatically generated feedback via reinforcement learning},
  author={Scarlatos, Alexander and Smith, Digory and Woodhead, Simon and Lan, Andrew},
  booktitle={International Conference on Artificial Intelligence in Education},
  pages={280--294},
  year={2024},
  organization={Springer}
}

@article{shao2024deepseekmath,
  title={Deepseekmath: Pushing the limits of mathematical reasoning in open language models},
  author={Shao, Zhihong and Wang, Peiyi and Zhu, Qihao and Xu, Runxin and Song, Junxiao and Bi, Xiao and Zhang, Haowei and Zhang, Mingchuan and Li, YK and Wu, Y and others},
  journal={arXiv preprint arXiv:2402.03300},
  year={2024}
}

@article{schulman2017proximal,
  title={Proximal policy optimization algorithms},
  author={Schulman, John and Wolski, Filip and Dhariwal, Prafulla and Radford, Alec and Klimov, Oleg},
  journal={arXiv preprint arXiv:1707.06347},
  year={2017}
}

@misc{qwen2025qwen25technicalreport,
      title={Qwen2.5 Technical Report}, 
      author={{Qwen} and An Yang and Baosong Yang and Beichen Zhang and Binyuan Hui and Bo Zheng and Bowen Yu and Chengyuan Li and Dayiheng Liu and Fei Huang and Haoran Wei and Huan Lin and Jian Yang and Jianhong Tu and Jianwei Zhang and Jianxin Yang and Jiaxi Yang and Jingren Zhou and Junyang Lin and Kai Dang and Keming Lu and Keqin Bao and Kexin Yang and Le Yu and Mei Li and Mingfeng Xue and Pei Zhang and Qin Zhu and Rui Men and Runji Lin and Tianhao Li and Tianyi Tang and Tingyu Xia and Xingzhang Ren and Xuancheng Ren and Yang Fan and Yang Su and Yichang Zhang and Yu Wan and Yuqiong Liu and Zeyu Cui and Zhenru Zhang and Zihan Qiu},
      year={2025},
      eprint={2412.15115},
      archivePrefix={arXiv},
      primaryClass={cs.CL},
      url={https://arxiv.org/abs/2412.15115}, 
}

@inproceedings{sonkar2024pedagogical,
  title={Pedagogical Alignment of Large Language Models},
  author={Sonkar, Shashank and Ni, Kangqi and Chaudhary, Sapana and Baraniuk, Richard},
  booktitle={Findings of the Association for Computational Linguistics: EMNLP 2024},
  pages={13641--13650},
  publisher={Association for Computational Linguistics},
  year={2024},
  doi={10.18653/v1/2024.findings-emnlp.797},
  url={https://aclanthology.org/2024.findings-emnlp.797/}
}

@inproceedings{zhang2025values,
  title={Edu-values: Towards evaluating the chinese education values of large language models},
  author={Zhang, Peiyi and Zhang, Yazhou and Wang, Bo and Rong, Lu and Tiwari, Prayag and Qin, Jing},
  booktitle={Companion Proceedings of the ACM on Web Conference 2025},
  pages={1519--1523},
  year={2025}
}

@article{xu2025edubench,
  title={Edubench: A comprehensive benchmarking dataset for evaluating large language models in diverse educational scenarios},
  author={Xu, Bin and Bai, Yu and Sun, Huashan and Lin, Yiguan and Liu, Siming and Liang, Xinyue and Li, Yaolin and Dong, Zhuangzhi and Zhang, Jingren and Deng, Yufan and others},
  journal={arXiv preprint arXiv:2505.16160},
  year={2025}
}

@misc{wei2025elmesautomatedframeworkevaluating,
      title={ELMES: An Automated Framework for Evaluating Large Language Models in Educational Scenarios}, 
      author={Shou'ang Wei and Xinyun Wang and Shuzhen Bi and Jian Chen and Ruijia Li and Bo Jiang and Xin Lin and Min Zhang and Yu Song and BingDong Li and Aimin Zhou and Hao Hao},
      year={2025},
      eprint={2507.22947},
      archivePrefix={arXiv},
      primaryClass={cs.CY},
      url={https://arxiv.org/abs/2507.22947}, 
}

@article{dan2023educhat,
  title={Educhat: A large-scale language model-based chatbot system for intelligent education},
  author={Dan, Yuhao and Lei, Zhikai and Gu, Yiyang and Li, Yong and Yin, Jianghao and Lin, Jiaju and Ye, Linhao and Tie, Zhiyan and Zhou, Yougen and Wang, Yilei and others},
  journal={arXiv preprint arXiv:2308.02773},
  year={2023}
}

@article{hayes2022practical,
  title={A practical guide to multi-objective reinforcement learning and planning},
  author={Hayes, Conor F and R{\u{a}}dulescu, Roxana and Bargiacchi, Eugenio and K{\"a}llstr{\"o}m, Johan and Macfarlane, Matthew and Reymond, Mathieu and Verstraeten, Timothy and Zintgraf, Luisa M and Dazeley, Richard and Heintz, Fredrik and others},
  journal={Autonomous Agents and Multi-Agent Systems},
  volume={36},
  number={1},
  pages={26},
  year={2022},
  publisher={Springer}
}

@techreport{learnlm2024improving,
  title={{LearnLM}: Improving {Gemini} for Learning},
  author={{LearnLM Team, Google}},
  institution={Google},
  year={2024},
  number={arXiv:2412.16429},
  url={https://arxiv.org/abs/2412.16429}
}
\bibliographystyle{iclr2027_conference}
\clearpage
\appendix
\section{Educational Reward Modeling Details}
\label{app:reward-modeling}
This section expands Figure~\ref{fig:framework}(a). We first specify the scenario-wise synthesis loop, then give the prompts for generation, evaluation, and refinement, followed by the \textbf{Edu-RM} scoring interface. The synthesis judge produces indicator-level supervision and rationales; \textbf{Edu-RM} subsequently predicts the three reward dimensions for policy rollouts.

\subsection{Iterative Data Synthesis Algorithm}
\label{app:synthesis-algorithm}
Algorithm~\ref{alg:synthesis} uses initial rubrics generated from expert interviews to guide iterative data synthesis. Each scenario's rubric assessments inform joint updates to its synthesis prompt and scoring standards, aiming to produce discriminative training examples for \textbf{Edu-RM}. Indicator names and H/P/C assignments remain fixed. GPT-5 performs generation, scoring, and refinement across 330 scenarios, with at most five rounds and two fixed anchors per scenario. The 12,390 examples pooled from accepted rounds retain different H/P/C score labels for learning judgments consistent with experts.

\begin{algorithm}[H]
\caption{Scenario-wise rubric-guided educational data synthesis}
\label{alg:synthesis}
\small
\begin{algorithmic}[1]
\REQUIRE 330 scenarios $\mathcal S$, learner profiles $\mathcal U$, expert-informed H/P/C criteria, maximum rounds $T=5$.
\REQUIRE GPT-5 for generation, scoring, and refinement; indicator weights and anchor drift-warning criteria.
\ENSURE Accepted labeled examples $\mathcal D_{\mathrm{syn}}$ and scenario-specific prompts and rubrics.
\STATE $\mathcal D_{\mathrm{syn}}\leftarrow\varnothing$.
\FOR{each scenario $s\in\mathcal S$}
\STATE Initialize $p_s^{(0)}$, $\mathcal R_s^{(0)}$, best score $q_s^*\leftarrow-\infty$, and two fixed anchors $\mathcal A_s$.
\FOR{$t=0,\ldots,T-1$}
\STATE Generate internal questions and matched user-facing questions using $p_s^{(t)}$ and profiles.
\STATE Generate answers to the internal questions; pair answers with their user-facing questions.
\STATE Score every pair against $\mathcal R_s^{(t)}$, obtaining indicator scores and rationales.
\IF{no usable generation/scoring output is available}
\STATE \textbf{break}.
\ENDIF
\STATE Average scores within each H/P/C indicator set using Equation~\eqref{eq:rubric}.
\STATE Compute $q_s^{(t)}$ as the indicator-weighted mean score over the current batch.
\STATE Rescore fixed question--answer anchors $\mathcal A_s$ to monitor rubric drift.
\STATE Issue a warning if anchor-score drift is detected.
\IF{$q_s^{(t)}>q_s^*$}
\STATE Save the current prompt and rubric; set $q_s^*\leftarrow q_s^{(t)}$; retain labeled examples in $\mathcal D_{\mathrm{syn}}$.
\ELSE
\STATE Restore the best prompt and rubric; \textbf{break}.
\ENDIF
\STATE Summarize indicator distributions, low-scoring answers, and scoring rationales as feedback.
\STATE Revise the generation prompt to address recurring response weaknesses.
\STATE Revise rubric descriptions and examples; preserve indicator identities and H/P/C assignments.
\STATE Use the resulting prompt and rubric in the next round.
\ENDFOR
\ENDFOR
\RETURN $\mathcal D_{\mathrm{syn}}$ and the best prompt--rubric state for each scenario.
\end{algorithmic}
\end{algorithm}

\paragraph{Generation state and two question forms.}
The internal question, \texttt{question\_for\_generation}, may include explicit length and formatting requirements. The user-facing \texttt{final\_question} expresses the same educational task, learner profile, and background in natural language. The generated answer is paired with the latter in the synthesized corpus. Keeping these two questions semantically aligned avoids losing the learner information needed to interpret a personalized answer when generation controls are removed.

\paragraph{Feedback and stability.}
The generation update uses low-scoring responses and their rationales to make instructions more specific. The rubric update uses score distributions and examples to clarify ambiguous criteria. Since both the response distribution and the rubric change, a higher score on a new batch alone cannot isolate response improvement. Two fixed anchors per scenario keep the question and answer unchanged while the rubric is rescored. Detected anchor-score drift triggers a warning. The batch score is the indicator-weighted mean over generated examples. A score strictly above the historical best saves the current prompt and rubric; a lower or tied score restores the best prompt and rubric together and stops the scenario's iteration. Anchor warnings do not determine acceptance or rollback. The maximum is five rounds, and accepted examples are pooled across rounds.

\subsection{Prompts for Data Generation and Iterative Refinement}
\label{app:synthesis-prompts}
The following English presentation templates consolidate the synthesis prompts from the earlier manuscript. Braced fields are filled with scenario-specific content. Indicator-level scores are integers from 0 to 5. Their within-dimension averages may be fractional and are rounded to integers before serialization as \textbf{Edu-RM} training targets.

\begingroup
\tcbset{eduprompt/.style={colback=black!2,colframe=black!35,colbacktitle=black!7,coltitle=black,boxrule=0.4pt,arc=1mm,left=7pt,right=7pt,top=6pt,bottom=6pt,fonttitle=\bfseries\small,fontupper=\small,before upper={\setlength{\parindent}{0pt}\setlength{\parskip}{4pt}\raggedright}}}
\newcommand{\promptfield}[1]{\texttt{\{#1\}}}
\begin{tcolorbox}[eduprompt,title={Data generation prompt}]
You are an expert in educational content design. Generate context-based educational question--answer material suited to the learner.

Educational scenario: \promptfield{scene}\par
Learner profile: \promptfield{persona\_section}\par
Current synthesis instructions: \promptfield{synthesis.prompt}\par
Number of examples: \promptfield{count}

For each example, return two questions:
\begin{enumerate}\setlength{\itemsep}{1pt}
\item \texttt{question\_for\_generation}: the internal question used to generate an answer. Include clear requirements for the desired length and detail.
\item \texttt{final\_question}: a natural user-facing question. Preserve the topic, background, role, learner characteristics, and constraints of the internal question. Express desired detail conversationally, without explicit word counts.
\end{enumerate}
Vary the opening: a learner or parent introduction, a request for an educational role, a direct request for help, or a description of a learning situation. Use realistic, conversational language. Avoid mechanical instructions and repeated question templates.

Return a JSON collection whose objects contain \texttt{question\_for\_generation} and \texttt{final\_question}.
\end{tcolorbox}

\begin{tcolorbox}[eduprompt,title={Indicator-level scoring prompt}]
You are an expert in educational quality assessment. Evaluate the answer using the specified indicator.

Educational scenario: \promptfield{scene}\par
Indicator name: \promptfield{metric\_name}\par
Description: \promptfield{metric\_description}\par
Positive example: \promptfield{positive\_example}\par
Negative example: \promptfield{negative\_example}\par
Question: \promptfield{question}\par
Answer: \promptfield{answer}

Assign an integer score from 0 to 5 using these anchors:\par
0: Fails the indicator entirely, contradicts its objective, or gives harmful or seriously misleading content relevant to it.\par
1: Meets almost none of the indicator's requirements.\par
2: Mostly does not meet the requirements.\par
3: Partially meets the requirements.\par
4: Meets the requirements well.\par
5: Fully meets the requirements.

Return JSON with \texttt{score} and \texttt{reasoning}. The reasoning should explain the rating with reference to the answer and the indicator.
\end{tcolorbox}

\clearpage
\begin{tcolorbox}[eduprompt,title={Generation-prompt refinement}]
You are an expert in improving prompts for educational data synthesis.

Educational scenario: \promptfield{scene}\par
Iteration: \promptfield{iteration}; prompt version: \promptfield{version}\par
Current synthesis prompt: \promptfield{currentprompt}\par
Average score: \promptfield{avg\_score}\par
Indicator-level summary: \promptfield{score\_summary}\par
Low-scoring examples and feedback: \promptfield{detailed\_scores}

Rewrite the synthesis prompt using the feedback:
\begin{enumerate}\setlength{\itemsep}{2pt}
\item Address low-scoring aspects with specific, actionable instructions.
\item Preserve realistic question wording, meaningful personalization, and responsible educational guidance.
\item Keep the two question forms consistent in meaning, background, role, and learner information. Use conversational detail cues in \texttt{final\_question}.
\item Retain effective instructions associated with higher-scoring aspects.
\item Remove ambiguous instructions and preserve the required JSON output fields: \texttt{question\_for\_generation} and \texttt{final\_question}.
\end{enumerate}
Return the revised synthesis prompt as plain text, without commentary or a JSON wrapper.
\end{tcolorbox}

\begin{tcolorbox}[eduprompt,title={Rubric refinement with fixed indicator identities}]
You are an expert in educational evaluation. Improve the clarity and discrimination of the current rubric.

Educational scenario: \promptfield{scene}\par
Iteration: \promptfield{iteration}; rubric version: \promptfield{version}\par
Current indicators: \promptfield{current\_metrics}\par
Score statistics, including any weighted summaries: \promptfield{score\_statistics}\par
Observed issues: \promptfield{observations}

Preserve the number of indicators and the exact \texttt{name}, \texttt{dimension}, and \texttt{weight} fields. Do not add, remove, or rename indicators. Revise only \texttt{description}, \texttt{positive\_example}, and \texttt{negative\_example}.

Inspect whether descriptions clearly distinguish quality levels and whether examples explain the scoring conditions. Use responses and scoring rationales to interpret apparently lenient, strict, or weakly discriminative indicators.

Return a JSON list in one-to-one correspondence with the input indicators. Each entry contains its unchanged name, dimension, and weight, together with the revised description and examples.
\end{tcolorbox}
\endgroup

\paragraph{From indicator feedback to reward labels.}
The scoring prompt is applied separately to each indicator. The resulting rationales support prompt and rubric refinement, while Equation~\eqref{eq:rubric} forms a three-dimensional label by averaging within each objective. The \texttt{weight} field in the inherited refinement template is preserved rubric metadata; it does not replace the within-dimension mean in Equation~\eqref{eq:rubric} or the equal coefficients of the policy's ideal-point loss.

\clearpage
\subsection{Reward Model Prompts and Score Interface}
\label{app:rm-prompts}
\textbf{Edu-RM} receives a question, a candidate response, and a unified scoring instruction. The H/P/C rubric blocks below are shared across all QA examples during reward-model training and policy-rollout scoring. Scenario-specific rubrics supply the training labels during synthesis, while these unified criteria define the reward model's scoring interface across contexts. The prompts are presented in English; H/P/C denotes the conceptual order used in the equations. The placeholders \texttt{\{H\_rubric\}}, \texttt{\{P\_rubric\}}, and \texttt{\{C\_rubric\}} are always replaced with the complete corresponding blocks below. The P block includes both cases; the scorer applies the case matching whether learner information is provided.

\begingroup
\tcbset{rmprompt/.style={colback=black!2,colframe=black!35,colbacktitle=black!7,coltitle=black,boxrule=0.4pt,arc=1mm,left=7pt,right=7pt,top=6pt,bottom=6pt,fonttitle=\bfseries\small,fontupper=\small,before upper={\setlength{\parindent}{0pt}\setlength{\parskip}{4pt}\raggedright}}}
\begin{tcolorbox}[rmprompt,title={\textbf{Edu-RM} system instruction}]
You are a multidimensional reward scoring model. Evaluate the response to the user's question along three dimensions: Helpfulness, Personalization, and Creativity. Assess each dimension using the rules below and return integer scores from 0 to 5.

\textbf{Helpfulness rules:} \texttt{\{H\_rubric\}}\par
\textbf{Personalization rules:} \texttt{\{P\_rubric\}}\par
\textbf{Creativity rules:} \texttt{\{C\_rubric\}}

Evaluate responsible guidance and factual reliability, then contextual adaptation and exploratory support. Creative or personalized content must remain appropriate to the user's request. Factual errors may affect more than one dimension when they make guidance misleading, infeasible, or poorly matched to the learner.

Return one JSON object containing only the three dimension scores. Do not include an explanation or other text.
\end{tcolorbox}

\begin{tcolorbox}[rmprompt,title={H: helpfulness}]
\textbf{Core meaning.} Constructive, rational, and inclusive attitudes; respect for others and diversity; honesty, lawfulness, civility, and courtesy; social responsibility and public-mindedness; opposition to extremism, discrimination, hatred, misinformation, and unlawful conduct.

In risk-related topics, be measured, cautious, and fact-based. Avoid incitement, labeling, and absolutist statements. Do not recommend dangerous or unverified practices. Suggest professional help when needed.

\textbf{Score anchors.}\par
5: Fully and clearly reflects these values, combining factual integrity with social responsibility. Proactively addresses risks and provides constructive value guidance in sensitive or risky situations. Language is measured, non-extreme, and respectful of diversity.\par
4: Clearly reflects these values, with appropriate language and no evident misconduct. Provides constructive guidance, but coverage is incomplete or potential risks are missed.\par
3: Generally neutral or mildly reflects these values. No evident misconduct, but value guidance is limited or necessary caution is missing.\par
2: Minor issues, such as inappropriate wording, overlooking limited risks, or a mildly disrespectful tendency. Remains broadly compliant.\par
1: Clearly neglects these values or shows irresponsibility, such as rash advice, indifference, or hostility toward opposing groups, without severe misconduct.\par
0: Violates these values through extremism, incitement to hatred or discrimination, unlawful conduct, clear misinformation, or encouragement of dangerous behavior.
\end{tcolorbox}

\clearpage
\begin{tcolorbox}[rmprompt,title={P: personalization}]
\textbf{Definition.} Assess how well the response adapts to the user's situation, available information, and constraints.

\textbf{Criteria.} Use stated age, role, preferences, knowledge, and resources appropriately. Match the task, tone, difficulty, and communication style to the context. Give feasible steps and priorities. Avoid inventing learner characteristics; use conditional suggestions or focused clarification when needed.

\textbf{When user-specific information is provided:}\par
5: Closely tailored to the profile and goals, with suitable difficulty, tone, resources, and executable guidance.\par
4: Generally well tailored, with minor generic details.\par
3: Some tailoring, but important needs or details are missed.\par
2: Mostly generic guidance with contextual mismatches.\par
1: Largely ignores the profile or offers poorly matched advice.\par
0: Contradicts the profile, relies on stereotypes, or gives risky advice because of poor fit.

\textbf{When user-specific information is absent:}\par
5: Context-appropriate guidance without unsupported assumptions; useful conditional advice or clarification.\par
4: Good contextual fit without unnecessary assumptions.\par
3: Reasonable but generic guidance with moderate fit.\par
2: Weak fit or unnecessary assumptions.\par
1: Invented user characteristics or advice mismatched to the situation.\par
0: Fabricated attributes or premises lead to inappropriate or risky guidance.
\end{tcolorbox}

\begin{tcolorbox}[rmprompt,title={C: creativity and exploratory support}]
\textbf{Definition.} Assess relevant novelty, diversity of ideas, and support for new and useful ways of thinking about the task.

\textbf{Criteria.} Consider alternative perspectives, approaches, and frameworks; questions, analogies, or thought experiments that encourage exploration; and a clear structure connecting goals, constraints, evaluation, and iteration. Concrete examples or steps should make the ideas usable.

\textbf{Score anchors.}\par
5: Multiple approaches, original insights, stimulating questions, and structured steps with examples or evaluation criteria.\par
4: Several perspectives or methods with clear structure and some originality.\par
3: Some alternatives or extensions, but largely conventional guidance.\par
2: Slight expansion beyond a direct answer; generic or weakly structured suggestions.\par
1: A direct answer with little exploratory support.\par
0: Superficial, formulaic, or off-topic content without useful exploration.

Novelty should serve the educational goal. A disconnected list of ideas without structure or usable criteria provides limited exploratory support.
\end{tcolorbox}
\endgroup

\paragraph{Serialization and dimensional order.}
The training target is a structured score sequence, optimized with the target-token cross-entropy objective in Equation~\eqref{eq:rm}. For each dimension, an averaged indicator score $\widehat r_k$ is converted to the integer target $\widetilde r_k=\lfloor\widehat r_k+0.5\rfloor$ before serialization, so half-integer values are rounded upward. This conversion yields targets in $\{0,1,\ldots,5\}$, matching the \textbf{Edu-RM} scoring instruction. Dimensional keys are mapped by meaning to construct $(r_H,r_P,r_C)$ from the generated scores.

\clearpage
\section{Training Details and Configuration}
\label{app:training}
\subsection{Reward Model Supervision}
\textbf{Edu-RM} and the educational policy have different supervised inputs and outputs. \textbf{Edu-RM} conditions on a question--response pair and the unified H/P/C rubric in Appendix~\ref{app:rm-prompts}, and predicts a structured three-score sequence. This scoring instruction is identical across QA examples. Their labels come from scenario-specific rubric scoring during data construction. Each reward-model training corpus contains 12,390 synthesized educational examples and 15,000 constructed low-quality negative examples, totaling 27,390 examples. In the before/after comparison, the before model uses iter~0 synthesis and the after model uses iterated synthesis. Both use the same set of 15,000 negative examples and identical training settings from Table~\ref{tab:rm-config}. The target-token cross-entropy in Equation~\eqref{eq:rm} trains this generative scoring interface. The indicator rationales support data refinement, while the score fields supply the reward targets.

Let $z$ denote the structured score sequence, with each dimensional average rounded to the nearest integer and half-integer ties rounded upward. With the unified scoring instruction implicit in the conditioning, the supervised objective is
\begin{equation}
\mathcal L_{\mathrm{RM}}(\phi)
=-\mathbb E_{(x,y,z)\sim\mathcal D_{\mathrm{RM}}}
\Big[\sum_{\tau=1}^{|z|}\log p_\phi(z_\tau\mid x,y,z_{<\tau})\Big],
\label{eq:rm}
\end{equation}
Here $\phi$ denotes the reward-model parameters, and the loss is computed over score-sequence tokens.

\paragraph{Negative-example construction.}
For the same questions used in synthesis, we use Qwen2.5-0.5B and Qwen2.5-3B to generate low-quality responses. We score these question--response pairs using the rubrics from the final iteration to obtain their H/P/C labels.

\paragraph{Policy training data.}
Policy SFT uses 5,000 question--response pairs selected from the iteratively synthesized data based on their combined helpfulness, personalization, and creativity scores. RL uses 3,880 synthesized prompts covering both educational and general questions.

\subsection{\textbf{IdealPO}: Policy Optimization Algorithm}
\label{app:algorithm}
Ideal-point Policy Optimization (\textbf{IdealPO}) is the policy optimization algorithm in \textbf{EduAlign}. It combines Ideal-Point Marginal Credit (\textbf{IMC}), which measures each response's contribution to reducing the group's ideal-point deficit, with Continuous Conflict Weighting (\textbf{CCW}), which attenuates credits with opposing dimensional contributions. Algorithm~\ref{alg:loo} gives the complete procedure.
\begin{algorithm}[htbp]
\caption{Ideal-point Policy Optimization (\textbf{IdealPO})}
\label{alg:loo}
\begin{algorithmic}[1]
\REQUIRE Policies $\pi_\theta$, $\pi_{\mathrm{old}}$, $\pi_{\mathrm{ref}}$; reward model $f_\phi$.
\REQUIRE Group size $n$, dimensions $K$, ceiling $M$, conflict strength $\lambda$, tolerance $\varepsilon$.
\FOR{each prompt $x$ in the rollout batch}
\STATE Sample $y_1,\ldots,y_n$; score independently: $r_{i,k}=f_\phi(x,y_i)_k$.
\STATE $\mu_k\leftarrow n^{-1}\sum_i r_{i,k}$; $\ell_k\leftarrow(1-\mu_k/M)^2$ for every $k$.
\FOR{$i=1,\ldots,n$}
\STATE $\mu_{-i,k}\leftarrow(n\mu_k-r_{i,k})/(n-1)$ for every $k$.
\STATE $c_{i,k}\leftarrow(1-\mu_{-i,k}/M)^2-\ell_k$; $b_i\leftarrow K^{-1}\sum_k c_{i,k}$.
\STATE $P_i\leftarrow K^{-1}\sum_k\max(c_{i,k},0)$; $N_i\leftarrow K^{-1}\sum_k\max(-c_{i,k},0)$.
\STATE $q_i\leftarrow[1-\lambda\,2\min(P_i,N_i)/(P_i+N_i+\varepsilon)]b_i$.
\ENDFOR
\STATE Compute $\bar q$ and population standard deviation $\sigma_q$ over this prompt's responses.
\STATE $A_i\leftarrow(q_i-\bar q)/\sigma_q$ if $\sigma_q>\varepsilon$; otherwise $A_i\leftarrow0$.
\ENDFOR
\STATE Broadcast $A_i$ to response tokens, without a second batch normalization.
\STATE Update $\theta$ with Equation~\eqref{eq:policy}.
\end{algorithmic}
\end{algorithm}

\subsection{Credit Derivation and Normalization Properties}
\label{app:credit-properties}
For $n>1$, the full and leave-one-out means are
\begin{equation}
\mu_k=\frac1n\sum_{j=1}^n r_{j,k},\qquad
\mu_{-i,k}=\frac{n\mu_k-r_{i,k}}{n-1}
=\mu_k-\frac{\delta_{i,k}}{n-1},
\label{eq:means}
\end{equation}
where
\begin{equation}
\delta_{i,k}=r_{i,k}-\mu_k,\qquad d_k=1-\frac{\mu_k}{M}.
\label{eq:deviation-deficit}
\end{equation}
The leave-one-out deficit is $d_k+\delta_{i,k}/[M(n-1)]$. Subtracting the full-group squared deficit yields
\begin{equation}
\begin{split}
c_{i,k}
&=\left(d_k+\frac{\delta_{i,k}}{M(n-1)}\right)^2-d_k^2\\
&=\frac{2d_k\delta_{i,k}}{M(n-1)}
+\frac{\delta_{i,k}^2}{M^2(n-1)^2}.
\end{split}
\end{equation}
For a fixed positive $\delta_{i,k}$, the derivative with respect to $d_k$ is $2\delta_{i,k}/[M(n-1)]>0$. This is the deficit sensitivity described in Section~\ref{sec:loo}: a given positive deviation receives more credit when the group mean is farther below the ceiling.

\paragraph{Dimensional signs and group dependence.}
On $[0,M]$, the squared deficit $(1-\mu/M)^2$ strictly decreases as $\mu$ increases. Consequently, $c_{i,k}$ has the same sign as $r_{i,k}-\mu_k$, and equals zero when the deviation is zero. The scalar credit satisfies $b_i=\mathcal L_{\mathrm{ideal}}(\boldsymbol\mu_{-i})-\mathcal L_{\mathrm{ideal}}(\boldsymbol\mu)$. Credits depend on the other responses sampled for the same prompt. Their deficit sensitivity describes the credit assignment rule; it does not impose a lower bound on any objective or guarantee simultaneous improvement after a policy update.

\paragraph{Conflict attenuation and final advantages.}
With $\lambda\in[0,1]$, the conflict ratio in Equation~\eqref{eq:conflict} gives $w_i\in[1-\lambda,1]$. For the experimental setting $\lambda=0.5$, $q_i=w_i b_i$ preserves the sign of $b_i$ and reduces its magnitude by at most half. This also makes a conflicting negative credit less negative. Response-dependent weights can change the ordering of credits. Subsequent centering compares $q_i$ with $\bar q$, so the sign of $A_i$ need not match the sign of $b_i$ or $q_i$, and attenuation of a raw credit need not reduce its normalized advantage magnitude.

The group statistics are $\bar q=n^{-1}\sum_i q_i$ and $\sigma_q^2=n^{-1}\sum_i(q_i-\bar q)^2$. If every response receives the same positive weight, normalization cancels that common factor whenever the variance remains above the tolerance. CCW therefore changes standardized advantages through variation in weights across responses, apart from the tolerance boundary. When $\sigma_q\le\varepsilon$, all advantages are zero and the group supplies no reward-driven surrogate gradient; the KL term can still contribute to the update.

\subsection{Token-Level Policy Objective}
\label{app:policy-objective}
We use these advantages in a clipped policy objective with a reference-policy KL penalty. Index prompt--response pairs in a minibatch by $u\in\mathcal B$, and let $A_u$ be the advantage computed in pair $u$'s prompt group. At response token $\tau$, define $h_{u,\tau}=(x_u,y_{u,<\tau})$ and the likelihood ratio $\rho_{u,\tau}(\theta)=\pi_\theta(y_{u,\tau}\mid h_{u,\tau})/\pi_{\mathrm{old}}(y_{u,\tau}\mid h_{u,\tau})$. With $T=\sum_{u\in\mathcal B}|y_u|$ valid response tokens, we maximize
\begin{equation}
\begin{split}
\mathcal J(\theta)=\frac1T\sum_{u\in\mathcal B}\sum_{\tau=1}^{|y_u|}\Big[
&\min\!\left(\rho_{u,\tau}A_u,
\operatorname{clip}(\rho_{u,\tau},1-\eta,1+\eta)A_u\right)\\
&-\beta\,\widehat D_{\mathrm{KL},u,\tau}(\pi_\theta\Vert\pi_{\mathrm{ref}})\Big].
\end{split}
\label{eq:policy}
\end{equation}
Here, $\widehat D_{\mathrm{KL},u,\tau}$ denotes the token-level KL penalty against the reference policy. 
\clearpage
\subsection{Training Parameters}
\label{app:config}
\subsubsection{Reward Model Training}
\textbf{Edu-RM} is initialized from Qwen2.5-32B-Instruct and trained for two epochs using the supervised objective in Equation~\eqref{eq:rm}. Table~\ref{tab:rm-config} lists its training hyperparameters.

\begin{table}[H]
\centering
\caption{\textbf{Edu-RM} supervised training configuration.}
\label{tab:rm-config}
\small
\begin{tabular}{ll}
\toprule
Setting & Value\\
\midrule
Backbone & Qwen2.5-32B-Instruct\\
Training epochs & 2\\
Learning rate & $10^{-5}$\\
Learning-rate warmup fraction & 0.05\\
Minimum learning rate & $10^{-6}$\\
Weight decay & 0.01\\
Maximum sequence length (tokens) & 32,768\\
Global batch size & 256\\
\bottomrule
\end{tabular}
\end{table}

\subsubsection{Policy Optimization}
Table~\ref{tab:config} summarizes the shared training hyperparameters used to optimize the 7B and 32B policies with \textbf{IdealPO}.

\begin{table}[H]
\centering
\caption{Shared \textbf{IdealPO} training hyperparameters.}
\label{tab:config}
\small
\begin{tabular}{ll}
\toprule
Setting & Value\\
\midrule
Learning rate & $10^{-6}$\\
Weight decay & 0.01\\
Global update batch size & 512\\
Prompts per rollout round & 128\\
Responses per prompt & 8\\
Policy clipping threshold & 0.2\\
KL coefficient & 0.001\\
Gradient clipping & 0.5\\
Rollout temperature & 1.0\\
Maximum prompt length (tokens) & 4,096\\
Maximum response length (tokens) & 8,192\\
Ideal reward $M$ & 5.0\\
Conflict strength $\lambda$ & 0.5\\
\bottomrule
\end{tabular}
\end{table}

\clearpage
\section{Additional Experimental Results}
\label{app:experiments}

\subsection{Case Studies}
\label{app:qualitative}
\begingroup
\definecolor{caseH}{RGB}{218,234,247}
\definecolor{caseP}{RGB}{222,238,218}
\definecolor{caseC}{RGB}{252,233,203}
\definecolor{caseBorder}{RGB}{170,178,184}
\definecolor{caseHeader}{RGB}{242,244,245}
\newcommand{\caseH}[1]{{\sethlcolor{caseH}\hl{#1}}}
\newcommand{\caseP}[1]{{\sethlcolor{caseP}\hl{#1}}}
\newcommand{\caseC}[1]{{\sethlcolor{caseC}\hl{#1}}}
\newcommand{\caseomit}{\unskip\ \textcolor{gray}{[\ldots]}\par}
\newcommand{\caselegend}{\noindent\small\colorbox{caseH}{\strut H: helpfulness}\quad\colorbox{caseP}{\strut P: personalization}\quad\colorbox{caseC}{\strut C: creativity}\par}
\tcbset{casepanel/.style={colback=white,colframe=caseBorder,colbacktitle=caseHeader,coltitle=black,boxrule=0.45pt,arc=1mm,left=6pt,right=6pt,top=5pt,bottom=5pt,fonttitle=\bfseries\small,fontupper=\fontsize{9}{10.5}\selectfont,before upper={\setlength{\parindent}{0pt}\setlength{\parskip}{2.5pt}\raggedright}},casequestion/.style={colback=caseHeader,colframe=caseBorder,boxrule=0.35pt,arc=1mm,left=6pt,right=6pt,top=4pt,bottom=4pt,fontupper=\fontsize{9}{10.5}\selectfont}}

\noindent\small
We compare 32B Instruct with \textbf{EduAlign}-32B, using sample~01 for both. These two examples were selected for higher Claude Sonnet 4.6 H/P/C ratings. Text is translated from Chinese; [\ldots] marks omissions. Author highlights indicate rubric dimensions, including contextual adaptation under P.
\par\medskip
\noindent\textbf{Case 1: Discussing a charitable-donation controversy}\hfill\textit{ID 172}
\begin{tcolorbox}[casequestion]
\textbf{Question.} I am a Grade 10 student. Recently, I saw a celebrity being attacked online over charitable donations. My classmates have different opinions, and I am confused. As a moral education teacher, could you explain the ethical issues involved? Please also suggest how we Grade 10 students can actively think about and exchange our views through group discussions or creative expression.
\end{tcolorbox}
\caselegend
\smallskip
\noindent
\begin{minipage}[t]{0.485\linewidth}
\begin{tcolorbox}[casepanel,title={Instruct},equal height group=caseonecompact]
\caseomit
\textbf{Cyberbullying and moral coercion.} Some negative online comments can develop into personal attacks or even cyberbullying. This raises questions about online moral norms: how to \caseH{respect others and avoid malicious comments} in online spaces.
\caseomit
\textbf{Suggestions for group discussion and creative expression}

\textbf{Group discussion:}

\caseP{You can organize group discussions around the ethical issues above}, encouraging students to consider different perspectives.

Ask students to share their views on celebrities' charitable donations and explain why they hold these views.

Discuss how to balance public figures' social responsibilities with their right to privacy.

\textbf{Creative expression:}

Create posters or leaflets summarizing the group's discussion and emphasizing the importance of viewing charitable actions positively.

\caseC{Write a short play or skit depicting a celebrity facing online attacks after making a donation, and propose solutions.}
\caseomit
\end{tcolorbox}
\end{minipage}\hfill
\begin{minipage}[t]{0.485\linewidth}
\begin{tcolorbox}[casepanel,title={\textbf{EduAlign}-32B},equal height group=caseonecompact]
\caseomit
\textbf{Freedom of expression and moral judgment.} Online commenters have the right to express different opinions, but \caseH{where is the boundary between legitimate criticism and malicious attacks?} What standards should guide moral judgment?
\caseomit
\caseP{Discussion framework and creative expression for Grade 10 students}
\caseomit
\textbf{Role-play (10 minutes):}

\caseC{Take the roles of the celebrity, beneficiaries, critics, and ordinary internet users, and explain each position.}

For example, the celebrity can emphasize intentions and challenges, while critics focus on a lack of transparency.
\caseomit
\textbf{Reflection and consensus (10 minutes):}

Review how feelings and perspectives changed during the discussion.

Try to reach a basic consensus on celebrity philanthropy, or at least \caseH{an agreement to respect diverse viewpoints}.
\caseomit
\textbf{Multimedia story map:}

\caseC{Create a digital or hand-drawn story map centered on the celebrity's donation, with branches showing different groups' perspectives and feelings} (such as beneficiaries, fans, and critics).
\caseomit
\textbf{Mediator.} \caseP{When discussion stalls, use neutral language such as} \caseH{``I notice that both sides care about X. Could you explain your concerns further?''} to ease tension.
\caseomit
\end{tcolorbox}
\end{minipage}

\clearpage
\noindent\textbf{Case 2: Learning mechanical efficiency with simulations}\hfill\textit{ID 121}
\begin{tcolorbox}[casequestion]
\textbf{Question.} I am a Grade 9 girl. I am introverted and enjoy thinking independently, but I rarely dare to speak in class. I recently used physics simulation software to experiment with the efficiency of simple machines, and I am unsure how simulations differ from real experiments. Could you explain their respective strengths and weaknesses for investigating and predicting physical phenomena? How can I use simulations to improve my understanding and prediction skills? I would like to explore this in depth.
\end{tcolorbox}
\caselegend
\smallskip
\noindent
\begin{minipage}[t]{0.485\linewidth}
\begin{tcolorbox}[casepanel,title={Instruct},equal height group=casetwocompact]
\caseomit
\textbf{Limitations.} Virtual experiments are usually \caseH{based on models and assumptions and may not fully reflect the complexity of the real world}.
\caseomit
\textbf{How can you use simulations to improve?}

\textbf{1. Understand the theory.} Before running simulations, make sure you have mastered the relevant physics theory. This will help you connect theory with practice during experiments.

\textbf{2. Comparative analysis.} Try designing experiments, run them virtually, and record the data. If possible, \caseC{verify them with real experiments, compare the differences, and identify their causes}.

\textbf{3. Adjust parameters.} Use the software's parameter controls to \caseC{change variables (such as mass or speed) and observe how these changes affect the results}. This will help deepen your understanding of physical laws.

\textbf{4. Discussion.} Share your simulation findings with classmates or teachers. Discussion can offer new perspectives and deepen your understanding of physical phenomena.

\textbf{5. Reflection.} After each simulation, review the process: what went well and what needs improvement? Gradually develop your experimental-design and data-analysis skills.
\caseomit
\end{tcolorbox}
\end{minipage}\hfill
\begin{minipage}[t]{0.485\linewidth}
\begin{tcolorbox}[casepanel,title={\textbf{EduAlign}-32B},equal height group=casetwocompact]
\caseomit
\textbf{Connect theory with practice:}

Before a virtual experiment, review the relevant physics, such as \caseP{the mechanical-efficiency formula}, $\eta=W_{\mathrm{useful}}/W_{\mathrm{total}}$. \caseC{Calculate the expected result on paper, then compare it with the simulation output.}

\caseC{Analyze why the two differ: is it due to the model's assumptions or a misunderstanding on your part?} Gradually narrow the gap between theory and practice.
\caseomit
\textbf{Hypothesis-driven learning:}

\caseC{Propose a hypothesis, such as ``Increasing the coefficient of friction reduces mechanical efficiency.''}

Test it with the software, gradually adjusting the coefficient of friction and observing the changes.

\caseC{Use the results to support or reject the hypothesis, then develop new conjectures and continue exploring.}
\caseomit
Virtual experiments are a powerful aid for exploring physics, but should not completely replace hands-on experience. Ideally, the two complement each other: first use simulations to build a conceptual framework, then \caseH{carry out real experiments safely under a teacher's guidance}, and finally apply the theory to open-ended problems.
\caseomit
Remember, \caseP{being introverted does not prevent deep thinking; everyone has their own learning pace}. Stay curious and enjoy scientific exploration!
\end{tcolorbox}
\end{minipage}
\endgroup

\clearpage
\subsection{Detailed Ablation Results}
\label{app:ablation-details}
Table~\ref{tab:ablation-deltas} compares \textbf{EduAlign} with the two 7B component ablations on the three tasks of each benchmark. AVG is the arithmetic mean of IP, PLS, and TMG for EduBench, and of ILPG, CQG, and KPE for ELMES.

\begin{table}[H]
\centering
\caption{Task-level 7B ablation results. $\Delta$ is ablation minus \textbf{EduAlign}; negative values indicate a lower score. AVG averages the three task scores, and differences are computed from values rounded to three decimal places.}
\label{tab:ablation-deltas}
\small
\begin{tabular}{llrrr}
\toprule
Benchmark & Task & \textbf{EduAlign} & $\Delta$ w/o \textbf{CCW} & $\Delta$ w/o \textbf{IMC}\\
\midrule
EduBench & IP & 8.245 & $-0.181$ & $-0.434$\\
 & PLS & 9.078 & $+0.002$ & $-0.106$\\
 & TMG & 9.230 & $-0.011$ & $-0.080$\\
 & AVG & 8.851 & $-0.063$ & $-0.207$\\
\midrule
ELMES & ILPG & 4.287 & $-0.047$ & $-0.347$\\
 & CQG & 4.000 & $-0.580$ & $-0.280$\\
 & KPE & 3.417 & $+0.000$ & $+0.103$\\
 & AVG & 3.901 & $-0.209$ & $-0.174$\\
\bottomrule
\end{tabular}
\end{table}

\paragraph{Effect of conflict weighting.}
Removing \textbf{CCW} lowers EduBench AVG by 0.063, mainly through the decrease in IP ($-0.181$). TMG decreases by 0.011, while PLS increases slightly by 0.002. On ELMES, AVG decreases by 0.209: CQG drops by 0.580 and ILPG by 0.047, while KPE is unchanged. The largest task-level effect of removing \textbf{CCW} is therefore on contextual question generation.

\paragraph{Effect of marginal credit.}
Removing \textbf{IMC} lowers all three EduBench task scores, with the largest decrease on IP ($-0.434$), and reduces AVG by 0.207. On ELMES, ILPG and CQG decrease by 0.347 and 0.280, respectively, while KPE increases by 0.103; AVG decreases by 0.174. \textbf{IMC} thus improves the task average on both benchmarks, with a trade-off on knowledge point explanation.

\FloatBarrier
\clearpage
\subsection{Fine-Grained Public-Benchmark Results}
\label{app:fullresults}
Tables~\ref{tab:fine-ip}--\ref{tab:fine-kpe} report task-specific rubric scores for Instruct, SFT, PPO, DAPO, GRPO, GDPO, and \textbf{EduAlign} at 7B and 32B. Bold denotes the highest score for each metric within a model scale, including ties.

\subsubsection{EduBench}
The breakdown covers idea provision (IP), personalized learning support (PLS), and teaching material generation (TMG). Each task uses its applicable rubric criteria, so the metric sets differ across tables.

\paragraph{Idea provision.}
\textbf{EduAlign} achieves the highest task score at both scales, with the highest scores on instruction following and domain knowledge (Table~\ref{tab:fine-ip}). DAPO scores higher on scenario integration and higher-order thinking at both scales.

\begin{table}[!htbp]
\centering
\caption{Fine-grained EduBench idea provision (IP) results at 7B and 32B.}
\label{tab:fine-ip}
\small
\setlength{\tabcolsep}{3pt}
\renewcommand{\arraystretch}{1.05}
\begin{tabular}{@{}>{\raggedright\arraybackslash}p{0.39\linewidth}rrrrrrr@{}}
\toprule
\multicolumn{8}{l}{\textbf{7B}}\\
Metric & Instruct & SFT & PPO & DAPO & GRPO & GDPO & \textbf{EduAlign}\\
\midrule
Task score & 6.867 & 6.740 & 7.400 & 7.819 & 7.834 & 7.845 & \textbf{8.245}\\
\midrule
Instruction Following \& Task Completion & 7.540 & 7.030 & 8.070 & 8.250 & 8.620 & 8.430 & \textbf{9.000}\\
Content Relevance \& Scope Control & 8.340 & 8.220 & 8.320 & 8.520 & 8.950 & 8.990 & \textbf{9.150}\\
Scenario Element Integration & 4.070 & 4.150 & 5.100 & \textbf{5.670} & 4.950 & 5.050 & 5.650\\
Basic Factual Accuracy & 8.130 & 7.830 & 8.040 & 8.570 & 8.600 & 8.890 & \textbf{8.990}\\
Domain Knowledge Accuracy & 7.180 & 7.350 & 7.820 & 8.090 & 8.330 & 8.120 & \textbf{8.720}\\
Reasoning Process Rigor & 6.870 & 6.900 & 7.860 & 8.150 & 8.250 & 8.260 & \textbf{8.600}\\
Clarity, Simplicity \& Inspiration & 7.270 & 7.140 & 7.610 & 7.830 & 8.130 & 8.140 & \textbf{8.490}\\
Higher-Order Thinking \& Skill Development & 5.540 & 5.300 & 6.380 & \textbf{7.470} & 6.840 & 6.880 & 7.360\\
\midrule
\multicolumn{8}{l}{\textbf{32B}}\\
Metric & Instruct & SFT & PPO & DAPO & GRPO & GDPO & \textbf{EduAlign}\\
\midrule
Task score & 7.985 & 8.094 & 7.820 & 8.321 & 8.234 & 8.275 & \textbf{8.547}\\
\midrule
Instruction Following \& Task Completion & 8.890 & 9.010 & 8.560 & 8.710 & 8.900 & 9.020 & \textbf{9.140}\\
Content Relevance \& Scope Control & 9.230 & 9.160 & 8.660 & 8.770 & 9.180 & 9.160 & \textbf{9.320}\\
Scenario Element Integration & 5.280 & 5.400 & 5.620 & \textbf{6.550} & 5.530 & 5.660 & 6.520\\
Basic Factual Accuracy & 9.170 & 9.100 & 8.660 & 9.000 & 9.010 & 9.110 & \textbf{9.220}\\
Domain Knowledge Accuracy & 8.570 & 8.530 & 8.140 & 8.640 & 8.790 & 8.640 & \textbf{9.050}\\
Reasoning Process Rigor & 8.570 & 8.660 & 8.230 & 8.600 & 8.690 & 8.700 & \textbf{8.970}\\
Clarity, Simplicity \& Inspiration & 8.090 & 8.380 & 7.960 & 8.430 & 8.460 & 8.450 & \textbf{8.570}\\
Higher-Order Thinking \& Skill Development & 6.080 & 6.510 & 6.730 & \textbf{7.870} & 7.310 & 7.460 & 7.590\\
\bottomrule
\end{tabular}
\end{table}
\FloatBarrier
\clearpage
\paragraph{Personalized learning support.}
\textbf{EduAlign} achieves the highest task score and the highest scores on instruction following and scenario integration at both scales (Table~\ref{tab:fine-pls}). Content relevance is slightly higher for GRPO at 7B and GDPO at 32B. At 32B, \textbf{EduAlign} ties SFT, DAPO, GRPO, and GDPO on personalization.

\begin{table}[!htbp]
\centering
\caption{Fine-grained EduBench personalized learning support (PLS) results at 7B and 32B.}
\label{tab:fine-pls}
\small
\setlength{\tabcolsep}{3pt}
\renewcommand{\arraystretch}{1.05}
\begin{tabular}{@{}>{\raggedright\arraybackslash}p{0.39\linewidth}rrrrrrr@{}}
\toprule
\multicolumn{8}{l}{\textbf{7B}}\\
Metric & Instruct & SFT & PPO & DAPO & GRPO & GDPO & \textbf{EduAlign}\\
\midrule
Task score & 8.510 & 8.530 & 8.694 & 8.868 & 9.028 & 8.998 & \textbf{9.078}\\
\midrule
Instruction Following \& Task Completion & 8.980 & 7.850 & 8.960 & 8.770 & 9.260 & 9.220 & \textbf{9.300}\\
Content Relevance \& Scope Control & 8.940 & 8.840 & 8.980 & 8.910 & \textbf{9.140} & 9.070 & 9.120\\
Scenario Element Integration & 8.240 & 8.430 & 8.460 & 8.840 & 8.910 & 8.860 & \textbf{9.000}\\
Personalization, Adaptation \& Learning Support & 8.590 & 8.870 & 8.770 & 8.990 & 8.960 & 8.990 & \textbf{9.000}\\
Higher-Order Thinking \& Skill Development & 7.800 & 8.660 & 8.300 & 8.830 & 8.870 & 8.850 & \textbf{8.970}\\
\midrule
\multicolumn{8}{l}{\textbf{32B}}\\
Metric & Instruct & SFT & PPO & DAPO & GRPO & GDPO & \textbf{EduAlign}\\
\midrule
Task score & 8.366 & 8.994 & 8.954 & 9.116 & 9.080 & 9.116 & \textbf{9.172}\\
\midrule
Instruction Following \& Task Completion & 8.950 & 9.190 & 9.430 & 9.450 & 9.420 & 9.510 & \textbf{9.740}\\
Content Relevance \& Scope Control & 8.980 & 9.120 & 9.070 & 9.160 & 9.110 & \textbf{9.170} & 9.150\\
Scenario Element Integration & 8.140 & 8.940 & 8.790 & 8.990 & 8.950 & 8.960 & \textbf{9.000}\\
Personalization, Adaptation \& Learning Support & 8.380 & \textbf{9.000} & 8.930 & \textbf{9.000} & \textbf{9.000} & \textbf{9.000} & \textbf{9.000}\\
Higher-Order Thinking \& Skill Development & 7.380 & 8.720 & 8.550 & \textbf{8.980} & 8.920 & 8.940 & 8.970\\
\bottomrule
\end{tabular}
\end{table}
\FloatBarrier
\clearpage
\paragraph{Teaching material generation.}
\textbf{EduAlign} achieves the highest task score at both scales and scores highest on instruction following and basic factual accuracy (Table~\ref{tab:fine-tmg}). At 32B, it also scores highest on content relevance and higher-order thinking, while GDPO scores slightly higher on role consistency and domain knowledge.

\begin{table}[!htbp]
\centering
\caption{Fine-grained EduBench teaching material generation (TMG) results at 7B and 32B.}
\label{tab:fine-tmg}
\small
\setlength{\tabcolsep}{3pt}
\renewcommand{\arraystretch}{1.05}
\begin{tabular}{@{}>{\raggedright\arraybackslash}p{0.39\linewidth}rrrrrrr@{}}
\toprule
\multicolumn{8}{l}{\textbf{7B}}\\
Metric & Instruct & SFT & PPO & DAPO & GRPO & GDPO & \textbf{EduAlign}\\
\midrule
Task score & 8.901 & 8.996 & 8.960 & 9.041 & 9.209 & 9.160 & \textbf{9.230}\\
\midrule
Instruction Following \& Task Completion & 9.580 & 9.390 & 9.530 & 9.370 & 9.670 & 9.540 & \textbf{9.790}\\
Role \& Tone Consistency & \textbf{9.000} & 8.980 & 8.980 & 8.970 & \textbf{9.000} & \textbf{9.000} & \textbf{9.000}\\
Content Relevance \& Scope Control & 9.430 & 9.340 & 9.310 & 9.330 & \textbf{9.680} & 9.520 & 9.630\\
Basic Factual Accuracy & 9.220 & 9.110 & 9.150 & 9.100 & 9.300 & 9.310 & \textbf{9.320}\\
Domain Knowledge Accuracy & 8.730 & 8.820 & 8.800 & 8.810 & \textbf{8.990} & 8.980 & \textbf{8.990}\\
Clarity, Simplicity \& Inspiration & 8.280 & 8.620 & 8.470 & 8.730 & 8.880 & 8.870 & \textbf{8.950}\\
Higher-Order Thinking \& Skill Development & 8.070 & 8.710 & 8.480 & \textbf{8.980} & 8.940 & 8.900 & 8.930\\
\midrule
\multicolumn{8}{l}{\textbf{32B}}\\
Metric & Instruct & SFT & PPO & DAPO & GRPO & GDPO & \textbf{EduAlign}\\
\midrule
Task score & 8.971 & 9.046 & 9.147 & 9.230 & 9.241 & 9.251 & \textbf{9.281}\\
\midrule
Instruction Following \& Task Completion & 9.730 & 9.770 & 9.820 & 9.810 & 9.810 & 9.810 & \textbf{9.910}\\
Role \& Tone Consistency & 8.960 & 9.000 & 9.000 & 9.000 & 8.990 & \textbf{9.020} & 9.000\\
Content Relevance \& Scope Control & 9.410 & 9.480 & 9.570 & 9.590 & 9.640 & 9.660 & \textbf{9.710}\\
Basic Factual Accuracy & 9.320 & 9.280 & 9.240 & 9.300 & 9.290 & 9.320 & \textbf{9.380}\\
Domain Knowledge Accuracy & 8.850 & 8.800 & 8.990 & 8.980 & 8.990 & \textbf{9.000} & 8.990\\
Clarity, Simplicity \& Inspiration & 8.250 & 8.360 & 8.690 & 8.950 & \textbf{8.980} & 8.960 & \textbf{8.980}\\
Higher-Order Thinking \& Skill Development & 8.280 & 8.630 & 8.720 & 8.980 & 8.990 & 8.990 & \textbf{9.000}\\
\bottomrule
\end{tabular}
\end{table}
\FloatBarrier
\clearpage
\subsubsection{ELMES}
Tables~\ref{tab:fine-ilpg}--\ref{tab:fine-kpe} report task scores and individual rubric criteria for ILPG, CQG, and KPE.

\paragraph{Interdisciplinary lesson plan generation.}
\textbf{EduAlign} achieves the highest task score and differentiated-support score at both scales (Table~\ref{tab:fine-ilpg}). At 32B, it scores highest on the assessment feedback loop and ties GRPO on core-knowledge coverage. Its teaching-condition feasibility score remains below the highest-scoring baseline at both scales.

\begin{table}[!htbp]
\centering
\caption{Fine-grained ELMES interdisciplinary lesson plan generation (ILPG) results at 7B and 32B.}
\label{tab:fine-ilpg}
\small
\setlength{\tabcolsep}{3pt}
\renewcommand{\arraystretch}{1.05}
\begin{tabular}{@{}>{\raggedright\arraybackslash}p{0.39\linewidth}rrrrrrr@{}}
\toprule
\multicolumn{8}{l}{\textbf{7B}}\\
Metric & Instruct & SFT & PPO & DAPO & GRPO & GDPO & \textbf{EduAlign}\\
\midrule
Task score & 3.313 & 3.800 & 3.700 & 4.010 & 4.107 & 3.980 & \textbf{4.287}\\
\midrule
Core-concept connections & 3.100 & 3.400 & 3.500 & 4.100 & 4.000 & 3.700 & \textbf{4.400}\\
Interdisciplinary logical integration & 2.800 & 3.400 & 3.500 & 4.100 & \textbf{4.200} & 3.700 & \textbf{4.200}\\
Core-knowledge coverage & 3.900 & 4.300 & 4.400 & 4.300 & 4.700 & 4.300 & \textbf{4.900}\\
Cognitive-conflict design & 1.700 & 2.200 & 2.200 & 2.400 & 2.400 & 2.300 & \textbf{2.900}\\
Context authenticity & 3.300 & 4.000 & 3.700 & 4.000 & 4.200 & 3.800 & \textbf{4.600}\\
Cognitive progression & 3.600 & 3.800 & 4.100 & 4.400 & 4.400 & 4.300 & \textbf{4.800}\\
Differentiated support & 1.800 & 2.600 & 2.200 & 2.800 & 2.800 & 3.000 & \textbf{3.500}\\
Student engagement & 4.000 & 4.500 & 4.800 & \textbf{5.000} & 4.900 & 4.800 & \textbf{5.000}\\
Process reflection records & 2.300 & 3.300 & 3.800 & 3.600 & 4.100 & 4.200 & \textbf{4.500}\\
Assessment feedback loop & 3.100 & 4.000 & 3.600 & 3.400 & 4.200 & \textbf{4.500} & \textbf{4.500}\\
Multidimensional performance assessment & 3.200 & 4.000 & 3.400 & 3.700 & \textbf{4.500} & 4.400 & \textbf{4.500}\\
Teaching-condition feasibility & 3.600 & 3.700 & 3.100 & \textbf{4.200} & 3.200 & 3.100 & 3.100\\
Suitability of materials and tools & 4.100 & 4.500 & 4.200 & 4.300 & 4.300 & 4.400 & \textbf{4.600}\\
Lesson-plan format compliance & 4.800 & \textbf{5.000} & 4.900 & \textbf{5.000} & \textbf{5.000} & \textbf{5.000} & 4.500\\
Content consistency & 4.400 & 4.300 & 4.100 & \textbf{4.900} & 4.700 & 4.200 & 4.300\\
\midrule
\multicolumn{8}{l}{\textbf{32B}}\\
Metric & Instruct & SFT & PPO & DAPO & GRPO & GDPO & \textbf{EduAlign}\\
\midrule
Task score & 3.520 & 4.100 & 3.910 & 4.327 & 4.487 & 4.433 & \textbf{4.513}\\
\midrule
Core-concept connections & 3.400 & 4.200 & 3.800 & 4.500 & 4.600 & \textbf{4.700} & \textbf{4.700}\\
Interdisciplinary logical integration & 3.300 & 3.900 & 3.700 & \textbf{4.800} & 4.600 & 4.600 & 4.700\\
Core-knowledge coverage & 4.200 & 4.700 & 4.600 & 4.800 & \textbf{5.000} & 4.600 & \textbf{5.000}\\
Cognitive-conflict design & 2.000 & 2.600 & 2.300 & 3.100 & \textbf{3.400} & 3.300 & \textbf{3.400}\\
Context authenticity & 3.500 & 4.400 & 3.700 & 4.700 & \textbf{4.900} & 4.600 & 4.600\\
Cognitive progression & 3.800 & 4.500 & 4.300 & 4.700 & \textbf{5.000} & 4.900 & 4.900\\
Differentiated support & 2.000 & 2.600 & 2.400 & 3.100 & 3.200 & 3.300 & \textbf{3.400}\\
Student engagement & 4.200 & 4.700 & 4.500 & \textbf{5.000} & \textbf{5.000} & 4.900 & \textbf{5.000}\\
Process reflection records & 2.400 & 3.300 & 3.700 & 4.000 & 4.100 & 4.200 & \textbf{4.300}\\
Assessment feedback loop & 3.100 & 4.100 & 3.800 & 4.300 & 4.600 & 4.600 & \textbf{4.700}\\
Multidimensional performance assessment & 3.200 & 4.200 & 4.100 & 4.300 & \textbf{5.000} & 4.700 & \textbf{5.000}\\
Teaching-condition feasibility & \textbf{4.100} & 3.900 & 3.800 & 3.600 & 3.400 & 3.800 & 3.500\\
Suitability of materials and tools & 4.400 & \textbf{4.700} & 4.600 & \textbf{4.700} & \textbf{4.700} & 4.400 & \textbf{4.700}\\
Lesson-plan format compliance & 4.800 & \textbf{5.000} & 4.800 & 4.700 & 4.900 & \textbf{5.000} & \textbf{5.000}\\
Content consistency & 4.400 & 4.700 & 4.500 & 4.600 & \textbf{4.900} & \textbf{4.900} & 4.800\\
\bottomrule
\end{tabular}
\end{table}
\FloatBarrier
\clearpage
\paragraph{Contextual question generation.}
At 7B, \textbf{EduAlign} achieves the highest task score and scores highest on question-stem quality, context quality, and teaching applicability, while PPO scores higher on solution-explanation quality (Table~\ref{tab:fine-cqg}). At 32B, \textbf{EduAlign} achieves the highest task score and the highest or tied-highest score on all five rubric criteria.

\begin{table}[!htbp]
\centering
\caption{Fine-grained ELMES contextual question generation (CQG) results at 7B and 32B.}
\label{tab:fine-cqg}
\small
\setlength{\tabcolsep}{3pt}
\renewcommand{\arraystretch}{1.05}
\begin{tabular}{@{}>{\raggedright\arraybackslash}p{0.39\linewidth}rrrrrrr@{}}
\toprule
\multicolumn{8}{l}{\textbf{7B}}\\
Metric & Instruct & SFT & PPO & DAPO & GRPO & GDPO & \textbf{EduAlign}\\
\midrule
Task score & 3.240 & 3.440 & 3.620 & 3.360 & 3.660 & 3.660 & \textbf{4.000}\\
\midrule
Question-stem quality & 3.100 & 3.400 & 3.000 & 3.000 & 3.100 & 3.000 & \textbf{3.600}\\
Context quality & 3.300 & 3.300 & 3.400 & 3.900 & 3.900 & 4.000 & \textbf{4.300}\\
Solution-explanation quality & 2.700 & 3.100 & \textbf{3.700} & 1.800 & 2.500 & 2.600 & 3.000\\
Teaching applicability & 3.000 & 3.300 & 3.200 & 3.300 & 3.800 & 3.800 & \textbf{4.200}\\
Value guidance & 4.100 & 4.100 & 4.800 & 4.800 & \textbf{5.000} & 4.900 & 4.900\\
\midrule
\multicolumn{8}{l}{\textbf{32B}}\\
Metric & Instruct & SFT & PPO & DAPO & GRPO & GDPO & \textbf{EduAlign}\\
\midrule
Task score & 3.680 & 4.040 & 4.220 & 3.960 & 4.100 & 4.160 & \textbf{4.640}\\
\midrule
Question-stem quality & 4.000 & 3.900 & 3.700 & 3.300 & 3.800 & 3.900 & \textbf{4.400}\\
Context quality & 3.600 & 4.100 & 4.200 & 4.100 & 4.400 & \textbf{4.500} & \textbf{4.500}\\
Solution-explanation quality & 3.300 & 3.800 & 4.200 & 3.500 & 3.200 & 3.300 & \textbf{4.500}\\
Teaching applicability & 3.400 & 3.700 & 4.200 & 4.000 & 4.100 & 4.100 & \textbf{4.800}\\
Value guidance & 4.100 & 4.700 & 4.800 & 4.900 & \textbf{5.000} & \textbf{5.000} & \textbf{5.000}\\
\bottomrule
\end{tabular}
\end{table}
\FloatBarrier
\clearpage
\paragraph{Knowledge point explanation.}
At 7B, \textbf{EduAlign} scores highest on emotional support, but its task score is lower than GDPO\textquotesingle{}s (Table~\ref{tab:fine-kpe}). At 32B, \textbf{EduAlign} achieves the highest task score and knowledge-point mastery score, and ties for the highest scores on teaching methods and content design. Its role-appropriateness score is lower than Instruct\textquotesingle{}s at both scales.

\begin{table}[!htbp]
\centering
\caption{Fine-grained ELMES knowledge point explanation (KPE) results at 7B and 32B.}
\label{tab:fine-kpe}
\small
\setlength{\tabcolsep}{3pt}
\renewcommand{\arraystretch}{1.05}
\begin{tabular}{@{}>{\raggedright\arraybackslash}p{0.39\linewidth}rrrrrrr@{}}
\toprule
\multicolumn{8}{l}{\textbf{7B}}\\
Metric & Instruct & SFT & PPO & DAPO & GRPO & GDPO & \textbf{EduAlign}\\
\midrule
Task score & 3.100 & 3.333 & 2.500 & 3.380 & 3.433 & \textbf{3.483} & 3.417\\
\midrule
Role appropriateness & \textbf{3.300} & 2.900 & 1.100 & 2.600 & 2.100 & 2.400 & 2.100\\
Emotional support & 3.500 & 4.300 & 2.300 & 4.400 & 4.800 & 4.700 & \textbf{4.900}\\
Knowledge-point mastery & 3.100 & 3.500 & \textbf{3.600} & 2.900 & 3.500 & 3.400 & 3.200\\
Appropriateness of teaching methods & 2.600 & 2.300 & 2.200 & 2.900 & 2.800 & \textbf{3.000} & 2.600\\
Appropriateness of content design & 3.600 & 3.600 & 3.400 & \textbf{3.900} & 3.700 & 3.400 & 3.800\\
Responsiveness to user characteristics & 2.500 & 3.400 & 2.400 & 3.600 & 3.700 & \textbf{4.000} & 3.900\\
\midrule
\multicolumn{8}{l}{\textbf{32B}}\\
Metric & Instruct & SFT & PPO & DAPO & GRPO & GDPO & \textbf{EduAlign}\\
\midrule
Task score & 3.633 & 3.217 & 3.120 & 3.800 & 3.733 & 3.633 & \textbf{3.933}\\
\midrule
Role appropriateness & \textbf{4.300} & 2.300 & 3.000 & 2.900 & 2.400 & 2.800 & 2.700\\
Emotional support & 4.000 & 3.500 & 3.400 & \textbf{5.000} & 4.700 & 4.700 & 4.700\\
Knowledge-point mastery & 3.800 & 4.200 & 3.500 & 3.700 & 4.100 & 4.100 & \textbf{4.400}\\
Appropriateness of teaching methods & \textbf{3.300} & 2.200 & 2.300 & 3.200 & 2.900 & 2.800 & \textbf{3.300}\\
Appropriateness of content design & 3.400 & 4.400 & 3.600 & 4.000 & \textbf{4.700} & 4.200 & \textbf{4.700}\\
Responsiveness to user characteristics & 3.000 & 2.700 & 2.900 & \textbf{4.000} & 3.600 & 3.200 & 3.800\\
\bottomrule
\end{tabular}
\end{table}
\FloatBarrier

\subsection{Expert and Matched Model-Judge Evaluation}
\label{app:human}
We collected questions through interviews with K--12 students and selected 200 to form the evaluation set shared by the experts and Claude. Table~\ref{tab:questionmix} reports the number of questions and topics covered in each category.

\begin{table}[H]
\centering
\caption{Question categories shared by expert and Claude assessment. Education-related questions account for 164 questions (82\%).}
\label{tab:questionmix}
\small
\begin{tabular}{@{}p{0.32\linewidth}rp{0.52\linewidth}@{}}
\toprule
Category & Questions & Content\\
\midrule
Subject knowledge and problem solving & 56 & Knowledge explanations and exercises in mathematics and other subjects.\\[3pt]
Teaching and learning support & 74 & Learning strategies, instructional design, educational scenarios, and guidance.\\[3pt]
Student development and peer issues & 6 & Emotions, psychological resilience, and classroom and friendship relationships.\\[3pt]
Open-ended inquiry and creation & 28 & Open-ended questions for writing, discussion, or project-based learning.\\
\midrule
General scenarios & 36 & General open-ended questions.\\
\midrule
Total & 200 & \\
\bottomrule
\end{tabular}
\end{table}

\FloatBarrier

\end{document}